\pdfoutput=1

\documentclass[11pt]{article}

\usepackage[final]{acl}
\usepackage{listings}

\usepackage{times}
\usepackage{latexsym}

\usepackage[T1]{fontenc}

\usepackage[utf8]{inputenc}

\usepackage{microtype}

\usepackage{inconsolata}

\usepackage{graphicx}
 \usepackage{pgf}
\usepackage{transparent}
\usepackage{array}
\usepackage{multirow}
\usepackage{multicol}
\usepackage{xprintlen}
\usepackage{csquotes}
\usepackage{enumitem}
\usepackage[normalem]{ulem}
\usepackage[export]{adjustbox}
\usepackage{float}
\usepackage{placeins}
\usepackage{subcaption}
\usepackage{makecell}
\usepackage{tipa} %
\usepackage{svg}
\usepackage{xurl}
\usepackage{bbding}
\usepackage[outline]{contour}
\usepackage{colortbl}

  \everymath=\expandafter{\the\everymath\displaystyle}
  \makeatletter\@ifpackageloaded{underscore}{}{\usepackage[strings]{underscore}}\makeatother

\newcolumntype{P}[1]{>{\centering\arraybackslash}m{#1}}

\definecolor{cellgrey}{RGB}{200, 200, 200}
\definecolor{celldgrey}{RGB}{150, 150, 150}

\contourlength{.1em}

\newcommand\columbiaC{{\normalfont 1}}
\newcommand\michiganM{{\normalfont 2}}
\newcommand\pennP{{\normalfont 3}}
\newcommand\knoxvilleT{{\normalfont 4}}

\newcommand\numcorpora{12}
\newcommand\numfilter{16}
\newcommand\numannots{ 1,054 }
\newcommand\minperc{0.007\%}
\newcommand\maxperc{0.18\%}
\newcommand\filtered{\textsc{C4.en}}
\newcommand\unfiltered{\textsc{C4.en.noBlocklist}}

\title{Data Caricatures: On the Representation of African American Language in Pretraining Corpora}

\author{
    \textbf{Nicholas Deas\textsuperscript{*\columbiaC}},
    \textbf{Blake Vente\textsuperscript{*\columbiaC}},
    \textbf{Amith Ananthram\textsuperscript{\columbiaC}},
    \textbf{Jessica A. Grieser\textsuperscript{\michiganM}},
    \\
    \textbf{Desmond Patton\textsuperscript{\pennP}},
    \textbf{Shana Kleiner\textsuperscript{\pennP}},
    \textbf{James Shepard\textsuperscript{\knoxvilleT}},
    \textbf{Kathleen McKeown\textsuperscript{\columbiaC}}
    \\
    \\
    \textsuperscript{\columbiaC} Columbia University, Department of Computer Science\\
    \textsuperscript{\michiganM} University of Michigan, Department of Linguistics\\
    \textsuperscript{\pennP} University of Pennsylvania, School of Social Policy and Practice, \\
    Annenberg School for Communications\\
    \textsuperscript{\knoxvilleT} University of Tennessee, Knoxville, Department of English
    \\
    \small{
       \{ndeas, kathy\}@cs.columbia.edu \quad rv2459@columbia.edu
     }
}

\begin{document}
\maketitle

\def\thefootnote{\arabic{footnote}}

\begin{abstract}
    
    With a combination of quantitative experiments, human judgments, and qualitative analyses,
    we evaluate 
    the quantity and quality of African American Language (AAL) representation in \numcorpora{} predominantly English, open-source pretraining corpora.
    We specifically focus on the sources, variation, and naturalness of included AAL texts representing the AAL-speaking community. 
    We find that AAL is underrepresented in all evaluated pretraining corpora compared to US demographics, constituting as few as \minperc{} and at most \maxperc{} of documents. We also find that more than 25\% of AAL texts 
    in C4 may be perceived as inappropriate for LLMs to generate and to reinforce harmful stereotypes. Finally, we find that most automated 
    filters are more likely to conserve White Mainstream English (WME) texts over AAL in pretraining corpora.~\footnote{We make our code available at \url{https://github.com/NickDeas/DataCaricatures}}
    
\end{abstract}

\section{Introduction}

    Recent work in NLP has become increasingly interested in evaluating and mitigating African American Language (AAL) biases in generative language models \cite{deas-aal, fleisig-chatgpt,hofmann-covert}. AAL, the comprehensive and rule-governed language variety used by many, but not all, and not exclusively, members of the US African American community \cite{grieser-dc}, 
    is known to be underrepresented in the C4 pretraining corpus \cite{dodge-c4}. In particular, C4 is largely composed of White Mainstream English (WME, see 
    \citet{baker-bell-aal}),
    the variety of English associated with White Americans and often found, for example, on Wikipedia.\footnote{While other works in linguistics and NLP use "African American Vernacular English (AAVE)" and "Standard American English (SAE)" among other terms, we use AAL and WME to highlight the inherent interaction between race and language in social hierarchies. Additionally, we typically use "language variety" to acknowledge the variation exhibited by AAL speakers across regions and contexts rather than a single, uniform dialect.}

    \begin{figure}
        \centering
        \includegraphics[width=0.9\linewidth,trim=1cm 0.7cm 1cm .7cm]{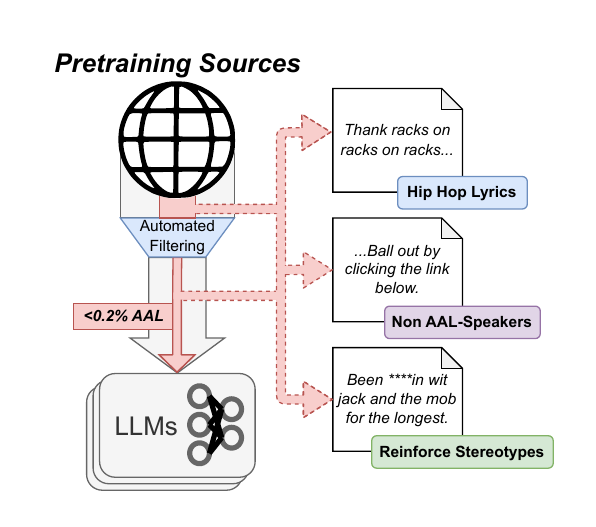}
        \caption{We evaluate the quantity and quality of automatically identified AAL documents in open source pretraining corpora. We find AAL is underrepresented across corpora and many documents contain texts that are misrepresentative of naturalistic AAL.
        }
        \label{fig:teaser}
    \end{figure}
    
    Given the close relationship between language and culture \cite{adilazuarda-culture}, the distributions of sources that texts are drawn from may also misrepresent the broader culture of communities when used to train models. For example, many websites publish hip hop lyrics which may be collected and incorporated into pretraining data. While hip hop 
    is an important component of Black culture and identity \cite{clay-hip-hop} as well as that of other communities, other registers of AAL (e.g., natural dialogue) are needed to more comprehensively capture the rich and diverse facets of Black culture.
    Furthermore, hip hop lyrics are not independently representative of AAL as a whole; lyrics do not necessarily reflect natural speech and no explicit indication of genre or register is provided during pretraining. 
    Prior studies validate this gap, finding that Black users perceive a lack of cultural knowledge in current language technologies; they often suspect that AAL speakers are overlooked in model development processes \cite{brewer-equitable, cunningham-understanding}. 
    
    Therefore, in addition to the \textit{quantity} of AAL that is represented, we argue that the \textit{quality} of its representation is also important.\footnote{Note, biases may also come from the technical representation of text in the form of text embeddings. This is distinct, however, from the representation that we explore in this work, which refers to how the language of AAL speakers is depicted in pretraining corpora.} 
    We define quality 
    as the extent to which texts authentically 
    portray the linguistic patterns--and consequently culture--of native AAL speakers. Accordingly, low quality representations of AAL (e.g., texts written by non-native speakers) pose representational harms to AAL speakers \cite{blodgett-harm}. 
    We focus on three guiding research questions: 
    \renewcommand\labelenumi{\textbf{RQ\theenumi)}}
    \setlist[enumerate,1]{leftmargin=3em}
    \begin{enumerate}
        \setlength\itemsep{-.2em}
        \item \textit{What percent of documents in modern, open-source pretraining corpora contain AAL and specific AAL features?}
        \item \textit{What is the quality of the included AAL texts in terms of diversity, authenticity, and lack of offensiveness?}
        \item \textit{How do modern data quality filtering strategies behave with AAL texts?} 
    \end{enumerate}
    \renewcommand\labelenumi{\theenumi}

    We use mixed-methods studies to address these questions, incorporating quantitative experiments, human judgments, and qualitative examination of the documents in pretraining corpora. 
    To investigate RQ1 (\autoref{sec:quantity}), we measure the prevalence of AAL in \textbf{\numcorpora{} open-source pretraining corpora} and find that \textbf{\minperc{} to \maxperc{} of documents among pretraining corpora we evaluate contain AAL}. In contrast, 80\% of African Americans ($\sim$10\% of Americans) are AAL speakers \cite{green-aae}. 
    To investigate RQ2 (\autoref{sec:quality}), we conduct automated and human evaluations of AAL texts in pretraining corpora and find that \textbf{a substantial number of documents are unrepresentative of naturalistic AAL}\footnote{By \textit{naturalistic AAL}, we refer to the linguistic patterns in text that most closely resemble everyday, spontaneous speech.} (e.g., perceived to be written by non-native speakers). 
    Surprisingly, some texts resembling AAL are actually posted by corporate social media accounts. 
    Finally, to investigate RQ3 (\autoref{sec:filtering}), we evaluate \textbf{\numfilter{} automated filtering approaches} 
    and find that\textbf{ most filters are more likely to conserve WME texts compared to AAL}, particularly for texts from social media or song lyrics.

\section{Background and Related Work}

    \textbf{AAL, Performativity, and Misrepresentation.}
        Research involving AAL in the NLP community has relied on texts drawn from different sources, such as speech transcripts \cite{farrington-coraal}
        and social media \cite{blodgett-variation}.
        Across such sources, AAL use can reflect varying degrees of what has been termed \textit{performative speech},
        language practices which use the rhetorical style of AAL such as \textit{signifyin'} \cite{claudia-signifyin} on a speaker's cultural background to construct or communicate Black identity. Because they primarily rely on language to communicate Black culture, performative register is especially prevalent in hip hop \cite{alim-roc} or social media language \cite{eisenstein-phon, ilbury-sassy}. 
        Recent work has also argued that the 
        linguistic patterns of these sources 
        may misrepresent
        the patterns of AAL speakers' when interacting with LLMs \cite{kleiner-camouflage}. 
        Alternatively, AAL can be misrepresented through the derogatory 
        use of features \cite{ronkin-mock}, diffusion of linguistic markers
        \cite{corradini-approp}, and 
        use of AAL-like language in marketing 
        \cite{gordon-comfort}. In this work, we examine the quality of AAL representation
        by considering performativity and these potential misrepresentations.
    
    \textbf{AAL Biases in Language Technologies.}
        The most prominent documentation and mitigation of AAL biases in text-based language models has 
        focused on toxicity detection (e.g., \citealt{sap-bias, harris-role, davidson-datasets,
        cheng-mitigate, halevy-mitigate}). More recent work 
        evaluates AAL biases in other contexts, including classification tasks
        through synthetic data \cite{ziems-value, dacon-nli}, summarization 
        \cite{keswani-summ, olabisi-diversity}, 
        stereotyping 
        behaviors \cite{fleisig-chatgpt, hofmann-covert}, generative language models \cite{groenwold-gen, deas-aal, deas-phon}, and reward models \cite{mire-rejected}. In contrast, 
        we specifically study
        AAL in pretraining data as a potential source of 
        bias.

    \textbf{Pretraining Data Quality.}
        \citet{hovy-sources} identify data as a fundamental source of bias,
        and prior work has attempted to trace various biases to 
        training data (e.g., \citet{ladhak-namenat, feng-political, dodge-c4}). Recent work has also increasingly been concerned with measuring and selecting high quality data for pretraining LLMs (e.g., \citet{wettig-qurating}). 
        Such studies,  
        however, use a variety of different 
        measures of data quality. A majority of prior work (e.g., \citealt{datacomp-lm}) primarily operationalizes quality through downstream model performance on broad benchmarks (e.g., \textsc{MMLU} \cite{hendrycks-mmlu} or \textsc{HellaSwag} \cite{zellers-hellaswag}).
        Other work measures quality using sources that are typically considered "high-quality" (e.g., Wikipedia) in developing automated data filters 
        (e.g., \textit{RedPajama}, \cite{redpajama}; \textit{The Pile}, \cite{thepile}). Finally, relatively fewer studies have examined intrinsic notions of quality,
        such as educational value or lack of harmful content \cite{wettig-qurating, sachdeva-askllm, gunasekar-textbooks}. With the aim of 
        identifying potential causes of AAL biases, we measure quality through a variety of means, focusing on texts' sources, 
        representativeness of naturalistic AAL, and lack of harmful reflections on AAL speakers.

\section{Data Extraction}

        \begin{table*}[!htbp]
            \centering
            \scriptsize
            \begin{tabular}{| r l |  P{2.5cm} P{1.8cm} P{1.5cm} r l |}
                \hline
                \multicolumn{2}{|c|}{\cellcolor{celldgrey}\textbf{Dataset}}                                   & \cellcolor{celldgrey}\textbf{Example Models}        & \cellcolor{celldgrey}\textbf{\# Docs (sample)}  & \cellcolor{celldgrey}\textbf{\% CommonCrawl Docs}    & \multicolumn{2}{c|}{\cellcolor{celldgrey}\textbf{\# AAL Docs (\%)}} \\
                \hline
                \multicolumn{7}{|c|}{\cellcolor{cellgrey}\textbf{Full Corpora}} \\
                \hline
                OpenWebText                     & \cite{gokaslan-openwebtext}   & SSDLM, RoBERTa        & 8M            & 0\%           & 858       & (.01\%) \\
                The Pile$^\dagger$              & \cite{thepile}                & GPT-Neo, GPT-J        & 140M          & 3\%           & 109,333   & (.08\%)  \\
                Dolmino (\textit{Dolmino-mix})  & \cite{olmo-2}                 & OLMo-v2               & 165M          & 83\%          & 53,937    & (.03\%)  \\
                \textbf{C4}                     & \cite{raffel-t5}              & T5                    & 365M          & 100\%         & 280,604   & (.07\%)  \\
                \textbf{C4.NoBlockList}         & \cite{dodge-c4}               & N/A                   & 395M          & 100\%         & 447,032   & (.11\%) \\
                RefinedWeb                      & \cite{refinedweb}             & FalconLM              & ~968M         & 100\%         & 1,143,244 & (.12\%)  \\
                \textbf{RedPajama}$^\dagger$    & \cite{redpajama}              & OpenLlama             & 968M          & 88\%          & 63,189    & (.007\%) \\
                \textit{FineWeb-Edu}            & \cite{fineweb}                & N/A                   & 1.8B          & 100\%         & 15,907    & (.0009\%) \\
                \textbf{Dolma}                  & \cite{soldaini-dolma}         & OLMo-v1               & 2.5B          & 78\%          & 3,119,429 & (.12\%)   \\
                \Xhline{1.5pt}
                \multicolumn{7}{|c|}{\cellcolor{cellgrey}\textbf{Sampled Corpora}} \\
                \hline
                Dolmino (\textit{Olmo-mix})     & \cite{olmo-2}                 & OLMo-v2               & 3.1B (238M)       & 96\%          & 317,178 & (.02\%$\pm$.001\%) \\
                DCLM-Baseline                   & \cite{datacomp-lm}            & DCLM-Baseline         & 3.2B (105M)       & 100\%         & 11,673 & (.01\%$\pm$.004\%) \\ %
                RedPajama-v2                    & \cite{redpajama-v2}           & N/A                   & 20.8B (176M)  & 100\%         & 684,099 & (.18\%$\pm$0.09) \\
                FineWeb                         & \cite{fineweb}                & N/A                   & ~48.6B (38M)  & 100\%         & 10,470 & (.03\%$\pm$.005\%) \\
                \hline
            \end{tabular}
            \caption{The \numcorpora{} open-source pretraining datasets evaluated, including example models pretrained on each dataset, size in raw unique documents, and proportion composed of Common Crawl. 
            $^\dagger$ indicates only non-copyrighted portions are available and included in analyses. \textbf{Bolded} corpora indicate that C4 is explicitly included in the corpus before further filtering. Given their size, the bottom 4 corpora are analyzed using a random sample of dataset shards; 99\% confidence intervals on each sampled corpus' estimate is included.
            }
            \label{tab:source-data}
        \end{table*}

        We examine the quantity and quality of AAL texts present in \numcorpora{} predominantly English, open-source pretraining corpora. All corpora are listed in \autoref{tab:source-data}.
        We consider FineWeb-Edu (italicized in \autoref{tab:source-data}) as a baseline, given that
        it strictly
        prioritizes highly educational 
        content (e.g., Wikipedia)
        over diverse everyday language use. 
        Following prior work's documentation of C4 \cite{dodge-c4} and because
        many corpora predominantly rely on Common Crawl texts, we focus our human judgments and AAL feature analyses on variants of the C4 corpus \cite{raffel-t5}.

    \subsection{Extracting AAL Subsets}

    To identify AAL texts, we follow \citeposs{dodge-c4} analysis of C4 as well as other related work (e.g., \citet{xia-demoting,sap-hatespeech,davidson-datasets}) and use a mixed-membership demographic-alignment classifier validated with common linguistic features of AAL
    \cite{blodgett-variation}.
    From eight full corpora, we extract all texts where AAL is the most likely classification (additional corpora details are included in \autoref{app:corpus-full}). For four exceedingly large corpora (each with more than 3 billion documents)--Dolmino (\textit{Olmo-mix}) \cite{olmo-2}, DCLM-baseline \cite{datacomp-lm}, FineWeb \cite{fineweb}, and RedPajama-v2 \cite{redpajama-v2}--we analyze a 250 GB sample of each corpus. Additional extraction methodology and corpus details are included in Appendices \ref{app:corpus-dets} and \ref{app:data-det}.~\footnote{While corpora like RedPajama-2 are intended to enable experimentation with automatic filtering rather than to be used as is for pretraining, we refer to all corpora as "pretraining corpora" for simplicity.}
    To account for AAL features that may be present within documents that largely reflect WME, we additionally extract a more conservative subset of texts using a threshold of 0.3 for further analyses.~\footnote{While prior work largely considers a threshold of 0.8 (e.g., \cite{xia-demoting}), we choose a threshold of 0.3 to calculate a more conservative estimate of feature prevalence while maintaining a manageable corpus size for analysis.
    Additional discussion is included in \autoref{app:threshold}.}

    We use the same block list of terms used in the original C4 corpus \cite{raffel-t5} to identify texts that would be filtered from the corpus before pretraining. This filtering yields two distinct variants: \unfiltered{} which does not employ the block list, and \filtered{} which lacks block list-identified texts \cite{dodge-c4}. 
    With these two variants, we conduct a fine-grained evaluation of the block list's impacts on AAL representation, prevalence of features, and human judgments of AAL texts.

\section{RQ1: How \textit{Much} is AAL Represented?}
    \label{sec:quantity}

    We first aim to quantify the prevalence of AAL in pretraining corpora. We conduct our analysis through both human judgments of documents as well as automated approaches. In this section, we first measure the proportion of AAL documents in each corpus and the count of overlapping AAL documents between corpora. We then consider the representation of individual AAL features.
    
    \subsection{Methods}

    \textbf{Human Judgments. }
        To complement automatic analyses, we collect human judgments of randomly sampled C4 texts within the subset of automatically extracted AAL texts.
        In sampling texts for judgments, we prioritize texts with higher probability of containing AAL according to the demographic alignment classifier as well as texts from \filtered{}. 
        
        We recruit three annotators to conduct the human judgments of the sampled texts; all annotators are self-reported native AAL speakers. Annotators also currently study or have previously studied linguistics or computational linguistics. Annotators are provided a text from either the AAL subset of \unfiltered{}{} or \filtered{}{} and asked to label each text first on two dimensions drawn from prior work \cite{deas-aal}:
        \textit{Human-Likeness} asks whether the text appears authored by a human and \textit{Linguistic Match} asks whether there are identifiable features of AAL. Both dimensions are judged on 4-point Likert scales (higher representing more human-like and more reflective of AAL). 
        We collect judgments for a total of \numannots texts. Annotators exhibit moderate to substantial agreement for binarized Human-Likeness and Linguistic Match dimensions ($\kappa = 0.581$ and $\kappa = 0.747$ respectively). Additional sampling, annotation, and interface details are included in \autoref{app:judgments}.
    
    \textbf{Automated Feature Extraction. }    
        To conduct a more fine-grained analysis, we estimate the distribution of morphosyntactic AAL features with automated methods. 
        Features are identified using the CGEdit model \cite{massis-cgedit}, a classifier developed with a human-in-the-loop framework. 
        The classifier considers 17 AAL morphosyntactic constructions, including features like habitual \textit{be} (e.g., "He be driving") and copula deletion (e.g., "He $\emptyset$ at home"). See \citealt{massis-cgedit} and \autoref{app:feat-explain} for a detailed list of features. As the classifier was shown to be reliable for examining distributions in large collections of text, we restrict analysis of features to trends in large data subsets rather than individual texts. 

    \subsection{Results}
        
    \textbf{AAL Frequency. }
        \autoref{tab:source-data} includes the proportion of documents labeled as AAL in each corpus evaluated. 
        Language understanding benchmarks, in part, drive pretraining data curation choices and are drawn from sources such as descriptive video captions (e.g., \textsc{HellaSwag} \cite{zellers-hellaswag}, \textsc{Swag} \cite{zellers-swag}) or academic exams (e.g., MMLU \cite{hendrycks-mmlu}). Therefore, we expect that the analyzed corpora contain few documents with AAL. 
        In line with this intuition, we find that AAL is extremely underrepresented across corpora used and intended for LLM pretraining, included in as few .007\% of documents (i.e., RedPajama).
        As expected, the FineWeb-Edu corpus, which employs among the strictest filtering approaches, exhibits the lowest percentage of documents containing AAL (.0009\%), while RedPajama-v2, which employs relatively limited filtering, exhibits the highest percentage (.18\%). Due to this trend and those found in prior work, we investigate filtering approaches further in \autoref{sec:filtering}.

        \begin{table}[!htbp]
            \centering
            \begin{tabular}{| P{2.5cm} P{1.7cm} | P{1.8cm} |}
                \hline
                \textbf{AAL Prob.} & \textbf{\# C4 Docs} & \textbf{\% AAL Judgments} \\
                \hline
                $0.5\leq p \leq 0.6$  & 41,930 & 44.7\% \\
                $0.6 \leq p \leq 0.7$ & 12,913 & 36.3\%\\
                $0.7 \leq p \leq 0.8$ & 4,319  & 36.7\% \\
                $0.8 \leq p \leq 0.9$ & 922    & 30.9\%\\
                $0.9 \leq p$          & 120    & 23.0\%\\
                \hline
            \end{tabular}
            \caption{Percent of texts labeled by human annotators as containing AAL features for each range of posterior probabilities of the demographic alignment classifier.
            }
            \label{tab:aal-freq}
        \end{table}

        To verify the automated analyses, \autoref{tab:aal-freq} presents the proportion of texts labeled by annotators as containing identifiable AAL features within different ranges of AAL posterior probability as output by the demographic alignment classifier.
        Annotators indicate that \textbf{a significant proportion of the texts 
        do not contain features of AAL} (55.3\% in the $0.5 \leq p \leq 0.6$ range). 
        We note that while the percentage of AAL judgments drops as AAL probability increases, this may be due to the classifier's sensitivity to spurious token-level features, such as abbreviations in search terms and artist names (e.g., "\textit{Lil} Wayne") that resemble lexical markers of AAL. To account for this, in our AAL feature and hip hop analyses, we consider all documents exceeding an AAL probability threshold of 0.3. See \autoref{app:class-exs} for further discussion and additional examples of classifier predictions.

\begin{figure}[htbp]
            \centering
            \includegraphics[width=0.95\columnwidth]{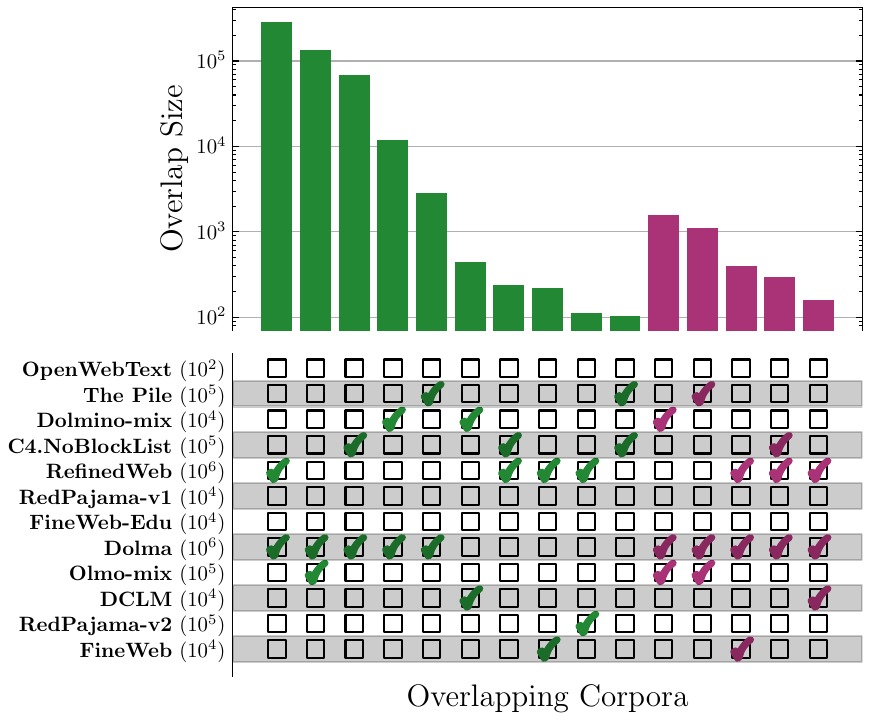}
            \caption{Counts of overlapping AAL documents among sets of corpora (indicated by checkmarks below the plot). Sizes of the AAL subsets for individual corpora are noted in parentheses. Overlapping sets smaller than 500 documents are not shown.
            }
            \label{fig:overlap}
        \end{figure}

    \textbf{AAL Document Overlap.}
        \autoref{fig:overlap} presents the count of overlapping AAL documents among 
        the corpora analyzed. While some corpora explicitly include others (e.g., Dolma is a superset of C4), these results suggest that there is substantial overlap in the AAL documents included for each corpus. Among all AAL documents considered, 17\% are duplicated in at least one other corpus. 
        Therefore, not only is AAL underrepresented, but there is also a lack of diversity across corpora.

\begin{figure}[!htbp]
            \centering
            \includegraphics[width=0.95\columnwidth]{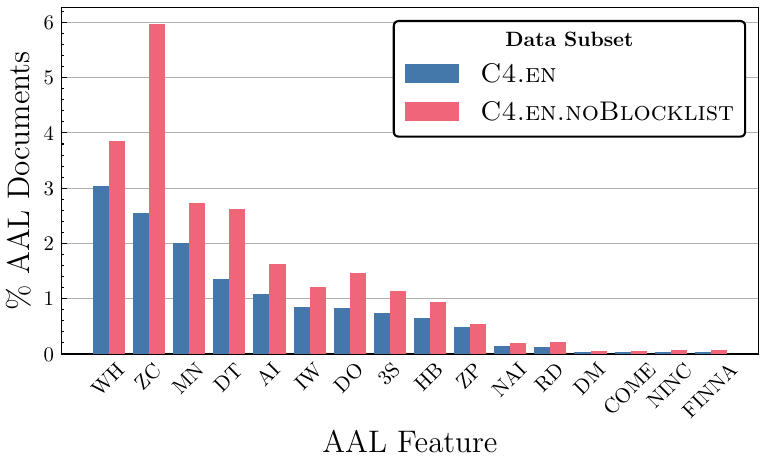}
            \caption{Document-level frequency 
            of morphosyntactic features among texts in \filtered{} and \unfiltered{}. Features ordered by frequency in \filtered{}. List of feature abbreviations and descriptions are included in \autoref{app:feat-explain}. 
            }
            \label{fig:filter-feats}
        \end{figure}

    \textbf{AAL Feature Frequency.}
        \autoref{fig:filter-feats} presents the frequency of each automatically detected AAL feature in \unfiltered{} and \filtered{}.
        Supporting \citeposs{dodge-c4} finding that the block list disproportionately filters AAL texts, we see that filtering lowers the frequency of all AAL features.
        Some of the most frequent features in 
        \unfiltered{}--zero copula (ZC) and multiple negatives (MN)--are among the most frequent documented features of naturalistic AAL. 
        Other features such as negative auxiliary inversion (NAI) and zero plurals (ZP), however, have extremely low representation despite being relatively common features among both urban and rural AAL speakers \cite{kortmann-ewave}. 
        
        While some features are represented in frequencies that generally reflect naturalistic AAL, 
        \autoref{fig:filter-feats} also illustrates how these frequencies are impacted by filtering. While the most frequent features are generally shared among both subsets, the order of most frequent features diverges. While Zero Copula (ZC; e.g., \textit{he} $\emptyset$ \textit{running}) is detected most often in \unfiltered{},
        it is far less represented in \filtered{}. 
        This alteration suggests that filtering may behave differently on AAL texts depending on the speaker's region, class, and other characteristics (e.g., Southern or working class; \citealt{wolfram2015regionality}).
        Not only does C4 filtering disproportionately remove AAL texts, but \textbf{filtering also systematically alters the distribution of AAL features.} 
        Feature analyses of the remaining corpora are included in \autoref{app:feature-results}, where we also observe substantial variation in feature distributions among different corpora.

\section{RQ2: How \textit{Well} is AAL Represented?}
\label{sec:quality}

    After characterizing the extent to which AAL is represented in open-source pretraining corpora, in this section, we aim to characterize the quality of the included AAL documents. To do so, we use the AAL subsets extracted in the previous experiments to further analyze the sources, naturalness, and inoffensiveness of the AAL texts.

    \subsection{Methods}
    
    \textbf{Human Judgments.}
        For each text that is judged to be human written and contain identifiable AAL features, annotators are asked to also rate each text on three additional dimensions. First, we follow \citet{deas-phon} and include \textit{Native Speaker} which asks whether the text appears to be naturally written by an AAL speaker. With the remaining two dimensions, we capture annotators' perceptions and linguistic attitudes toward a given text considering its inclusion in pretraining data.
        Inspired by \citet{fleisig-chatgpt}, \textit{Stereotype} first asks whether the text portrays a stereotypical or harmful representation of AAL or its speakers.
        Additionally, we include \textit{Appropriateness} which asks whether an annotator perceives a given text to be appropriate for language models to generate, and therefore, appropriate to include in pretraining data.
        Because we are interested in capturing the annotators' own perspectives on pretraining texts and their interpretations of these dimensions, we avoid providing detailed definitions of \textit{Stereotype} and \textit{Appropriateness} dimensions in particular,  motivated by sociolinguistics work on language attitudes and perceptions \cite{campbell-judge,labov-monitor}.
        
        All dimensions are again judged on 4-point Likert scales with higher values representing that a text is judged to likely have been written by a native speaker, to perpetuate stereotypes, and to be appropriate for an LLM to generate. Annotators exhibit substantial agreement for \textit{Native Speaker} ($\kappa$ = 0.619) although annotators' varying personal perspectives on the \textit{Appropriateness} and \textit{Stereotype} dimensions yield little to no agreement ($\kappa = $ 0.188 and -0.021 respectively). Given that these dimensions are highly dependent on individuals' own perceptions, this level of agreement is expected. Additionally, these results highlight the need to ensure that the inclusion of AAL speakers' perspectives captures
        the variety of perspectives.
        Additional details and discussion of agreement are included in \autoref{app:judgments}.

    \textbf{Misrepresentative Language.}
        To explore whether the representation of AAL in C4 is indicative of naturalistic AAL, we first estimate the presence of language resembling hip hop and rap lyrics in C4 variants. We particularly focus on song lyrics because 
        large-scale corpora of published hip hop lyrics are available, enabling the measurement of overlap with each pretraining corpus. Leveraging the aforementioned human judgments, we also estimate the perceived presence of AAL use by non-AAL speakers.
        
        We follow the deduplication approach of \citealt{brown-gpt-3} which measures the overlap of 8-13 token n-grams. While an ideal analysis would measure exact matches, through manual examination, we find instances of hip hop lyrics embedded in non-lyric text, censored terms, and orthographic differences (e.g., "going" vs. "goin") that would not be captured by such an approach. 
        We note that while we use a large corpus of song lyrics, a comprehensive dataset remains infeasible and our estimates likely reflect lower bounds. Methodological details are included in \autoref{app:hip-hop-method}.

    \subsection{Results}

        \begin{figure}
            \centering
            \includegraphics[width=0.95\columnwidth]{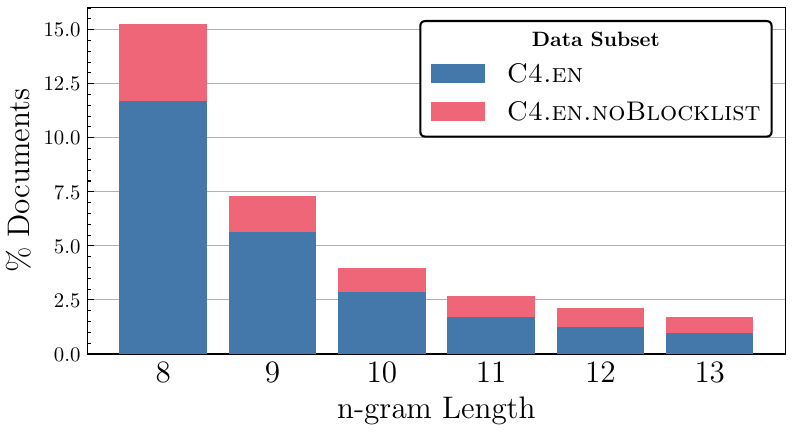} 
            \caption{Percent of AAL documents extracted from C4 containing shared n-grams with hip hop lyrics for varying n-gram token lengths. Tokenization uses the Llama-2 tokenizer. 
            }
            \label{fig:hip-hop-ngram}
        \end{figure}

        \begin{table}[!htbp]
            \centering
            \scriptsize
            \begin{tabular}{| P{2.2cm} | P{4.5cm} |}
                \hline
                C4 Subset & Text \\
                \hline
                \multirow{2}{*}{\centering \texttt{EN}} & \textit{You go dey hear: ha ha! Catch am, catch am! Thief, thief, thief! Catch am, catch am! Rogue, rogue, rogue!}  \\
                \cline{2-2}
                & \textit{Cant Say My Name But Rap About A N*ggas Wife. You So Black \& White Tryna Live A N*ggas Life... You Aint Wettin Nobody You Canady Dry} \\
                \hline
                \texttt{EN.NoBlockList} & \textit{Thank Racks on racks on racks on racks / None of my cars ain’t rented, all mine black, my windows tinted} \\
                \hline
            \end{tabular}
            \caption{Examples of hip hop lyrics found in \filtered{} and \unfiltered{}. Texts are shown exactly as they appear in the corpus.}
            \label{tab:hip-hop-ex}
        \end{table}

        \textbf{Hip Hop Lyrics.} \autoref{fig:hip-hop-ngram} presents the percentage of documents overlapping with hip hop lyrics by n-gram token length. Nearly 15\% of \unfiltered{} and 12\% of \filtered{} documents 
        overlap with hip hop lyrics (8-grams).
        Examples of lyrics identified by annotators are included in \autoref{tab:hip-hop-ex}. 
        While there are AAL features present, such as the use of negative concord in \textit{None of my cars ain't rented}, any inclusion of hip hop lyrics may be misrepresentative of how AAL speakers would interact with LLMs. Analyses of the remaining corpora are included in \autoref{app:all-hip-hop}, and additional related analyses are included in \autoref{app:add-hip-hop}. 

        \textbf{Non-AAL-Speakers.}
        AAL may also be misrepresented through use by non-AAL speakers, predominantly online. While it is infeasible to automatically determine the linguistic background of a text's author, among the \filtered{} texts annotators judged to contain AAL features, 51\% were judged unlikely to have been written by a native AAL speaker.
        Most of these texts (70\%) were perceived as not human-written.~\footnote{As an approximate reference, only 21\% of human-written AAL texts in \citet{deas-aal} were judged to be likely not human-written or unnatural.} The remaining texts, however, include corporate social media accounts adapting AAL-like language
        (e.g., 
        \textit{...this will get you where you need to be. Ball out by clicking the link below.}) as well as online forum posts (e.g. \textit{Dat be that real deal Hip Hop! ...Put some spect on it!}). 
        These texts often exaggerate the use of AAL features, contributing to misunderstandings of AAL speakers.
        We conclude that a \textbf{substantial portion of texts with AAL features are unlikely to have been written by an individual AAL speaker, and misrepresent naturalistic AAL}. 
        
        Many features of AAL are shared with other language varieties and the appearance of shared features in different linguistic contexts may complicate LLMs' ability to model these varieties during pretraining. Annotators noted many of these cases, such as the Hawai'ian 
        Creole text "\textit{...An make us shame cuz we no mo husban. Dat Yahweh make come up from da king ohana...}" which may be misinterpreted as Digital AAL \cite{cunningham-daal}.~\footnote{See \autoref{app:non-aal} for further discussion.}

        \begin{figure}[!htbp]
            \centering
            \includegraphics[width=0.95\columnwidth]{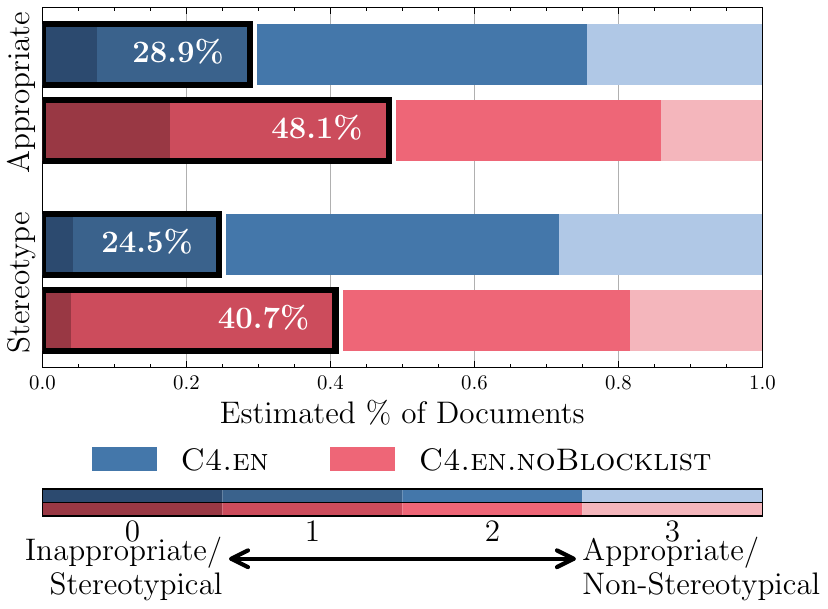}
            \caption{Distribution of annotator ratings on stereotype and appropriateness dimensions. Shades correspond to Likert scale values. Percentages shown in white reflect the proportion of texts judged to be inappropriate and to reinforce stereotypes (outlined in black).
            } 
            \label{fig:approp-stereo}
        \end{figure}

        \begin{table}[!htbp]
            \centering
            \scriptsize
            \begin{tabular}{| P{5cm} | P{0.5cm} P{0.5cm} |}
                \hline
                Text & A$\uparrow$ & S$\downarrow$ \\
                \hline
                \textit{...Been ****in wit jack and the mob for the longest. been ****in wit this MOB muzic 4 a very long time , Jack!..."} & 0 & 2 \\
                \hline
                \textit{He gon fuk around \& drown off this ?} & 0 & 0 \\
                \hline
                \textit{Drake is two levels up and Meek Mill is still not off the mark. The Toronto rapper just dropped another diss song “Back To Back” where he freestyle about his former friend and now nemesis.} & 3 & 1 \\
                \hline
            \end{tabular}
            \caption{Examples of AAL texts from \filtered{} with appropriateness (A) and stereotype (S) judgments.}
            \label{tab:approp-ex}
        \end{table}

        \textbf{Innapropriateness and Stereotyping.} \textit{Appropriateness} and \textit{Stereotype} judgment results are shown in \autoref{fig:approp-stereo}. 
        First, we find that the filtering applied to \unfiltered{} removes many inappropriate and stereotype-reinforcing texts as intended, with \unfiltered{} exceeding the rates of \filtered{}. We also find, however, that \textbf{substantial proportions of remaining texts in \filtered{} are judged to be inappropriate and stereotype-reinforcing}--28.9\% and 24.5\% respectively. Many of these texts contain variants of C4 block list terms or terms self-censored with asterisks. For example, the first two texts in \autoref{tab:approp-ex} are judged to be inappropriate for models to generate and contain references to block list terms (\textit{****in} and \textit{fuk}).

\section{RQ3: How does filtering impact AAL representation?}
    \label{sec:filtering}
    
    The previous experiments show that AAL is often underrepresented and misrepresented in modern, open-source pretraining corpora. We follow these experiments with an evaluation of automated filters to better understand potential underlying determinants of AAL representation.

    \subsection{Methods}
        
        \begin{table}[!htbp]
            \centering
            \small
            \begin{tabular}{| c  c | c c |}
                \hline
                Source                          & Dialect   & \# Documents              & Avg. Length \\
                \hline
                \multirow{2}{*}{RedPajama-v2}   & AAL       & \multirow{2}{*}{235,490}  & 509.8 \\
                                                & WME       &                           & 784.7 \\
                \Xhline{1.5pt}
                \multirow{1}{*}{Dialogues}      & AAL       & \multirow{1}{*}{11,787}   & 59.9 \\
                \hline
                \multirow{1}{*}{Song Lyrics}    & AAL       & \multirow{1}{*}{10,000}   & 441.0 \\
                \hline
                \multirow{1}{*}{Social Media}   & AAL       & \multirow{1}{*}{50,000}   & 19.7 \\
                \hline
            \end{tabular}
            \caption{Summary of filtering experiment datasets. 
            }
            \label{tab:filter-data}
        \end{table}

        The C4 block list is known to disproportionately filter AAL 
        from pretraining data \cite{dodge-c4}. To investigate how recent, model-based language, toxicity, and quality filters 
        affect AAL representation, we evaluate \numfilter{} automated filters on AAL texts.
        A full list with descriptions of each filter evaluated is included in \autoref{app:filters}.

        We conduct two sets of experiments to evaluate the behavior of automated filters on AAL texts. First, we evaluate automated filters on AAL and WME texts in RedPajama-v2 \cite{redpajama-v2} to examine how filters treat web texts that naturally occur in sources of pretraining data. 
        While this experiment reflects real use cases of filters, it does not provide insights into behavior across varying sources of AAL texts. 
        In a second controlled experiment, we collect AAL from different sources (dialogues, hip hop lyrics, social media) to evaluate how filters' behaviors differ among domains. 

        \textbf{RedPajama Filtering} RedPajama-v2 \cite{redpajama-v2} is a large pool of Common Crawl text with pre-computed filtering heuristics. 
            To evaluate automated filters, we extract all texts with an AAL posterior probability $\geq 0.8$ and randomly sample an equal number of texts with a White-aligned English probability of $\geq 0.8$ following prior work \cite{xia-demoting}.

    \def\filtplotwidth{1.0\textwidth}
    \begin{figure*}[htbp]
        \centering
        \begin{subfigure}[b]{\filtplotwidth}
            \centering
            \includegraphics[width=0.9\linewidth]{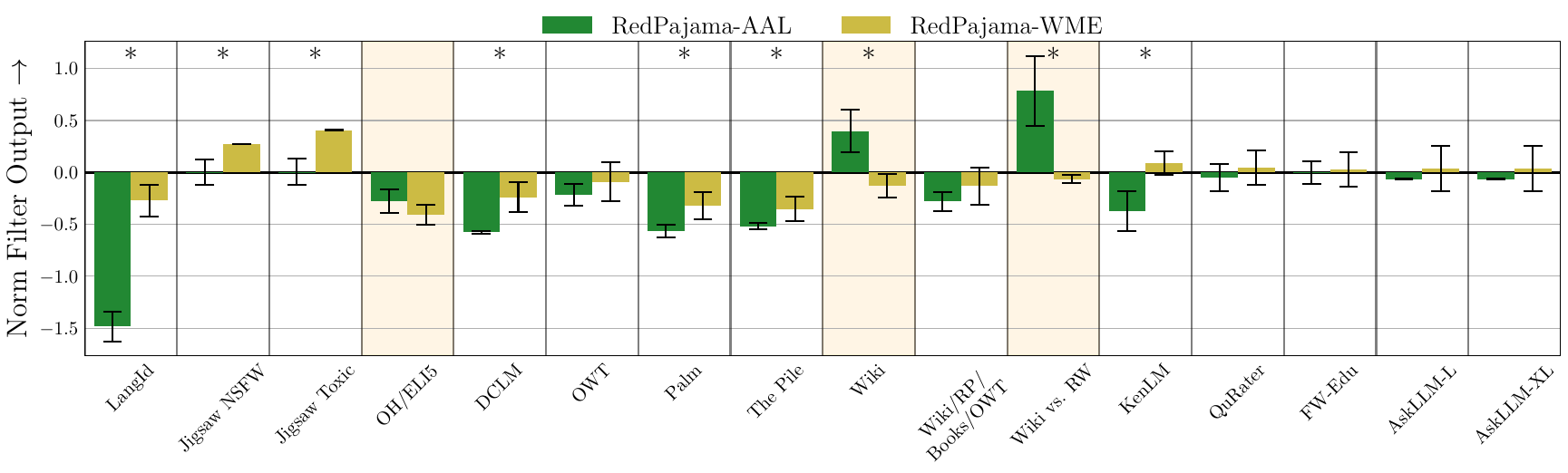}
        \end{subfigure}

        \begin{subfigure}[b]{\filtplotwidth}
            \centering
            \includegraphics[width=0.9\linewidth]{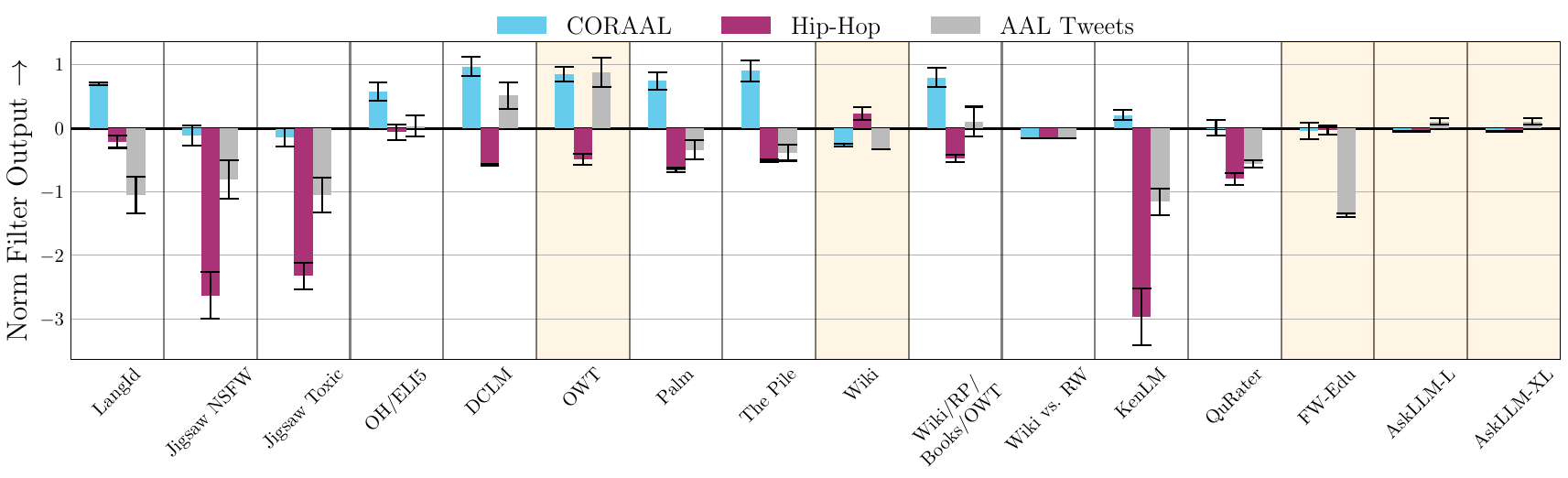}
        \end{subfigure}
        \caption{Average outputs of automated filters between AAL and WME RedPajama texts (top) and across AAL data sources (bottom). Outputs are standardized with larger values indicating texts are more likely to be preserved in the corpus. Error bars represent 99\% confidence intervals, and $^*$ (top) indicates significant differences ($p<0.01$). Shaded columns indicate filters that prefer AAL texts (top) and prefer Hip Hop/AAL Tweets (bottom).
        }
        \label{fig:filter-res}
    \end{figure*}

        \textbf{AAL Source Filtering.}
            To analyze different sources of AAL, we collect AAL texts from different domains 
            (as in \citet{deas-aal}).
            We evaluate automated filtering on three different domains of AAL-like language: social media, hip hop lyrics, and dialogues. 
            For \textit{social media} texts, we use the African American-aligned subset
            of the TwitterAAE corpus \cite{blodgett-variation}. For \textit{hip hop lyrics} we randomly sample full song lyrics from hip hop and rap songs.
            Finally, for \textit{dialogues}, we use CORAAL \cite{farrington-coraal} to represent naturalistic AAL. \footnote{Consecutive turns by the same speaker are merged to account for dialogue turns consisting of few tokens (e.g., "Yes").}
            A summary of all filter evaluation datasets can be found in \autoref{tab:filter-data}.

        \textbf{Filter Evaluation.}
            We run each filter on the aforementioned datasets for evaluation. We compare the z-score normalized ($z = \frac{x - mean}{stdev}$) predictions for each automated filter based on its form: probabilities for classifiers, perplexities for n-gram models, and ratings for LLM-as-a-judge approaches. For all classifiers and rating-based models, we consider the score given to the \textit{positive} class or direction such that higher scores represent a higher likelihood of text being preserved. We negate perplexities for applicable filters such that higher z-scores similarly indicate a higher likelihood of being conserved. We assess significant differences using two-tailed t-tests of the means.  

    \subsection{Results}

    \textbf{RedPajama-v2.} We first examine how automated filters behave with AAL and WME data distributions similar to those in pretraining corpora using RedPajama-v2 \cite{redpajama-v2}. \autoref{fig:filter-res} (top) presents the normed average filter outputs for the positive label (i.e., the label more likely to conserve a given text) on the AAL and WME subsets of RedPajama-v2. \textbf{Most filters (13 of \numfilter{}) are more likely to remove AAL texts than WME texts}. Supporting prior work, language \cite{blodgett-racial} and toxicity \cite{sap-hatespeech} filters are included among those that assign lower scores to AAL texts. Interestingly, two of the remaining filters (Wiki and Wiki vs. RW) include Wikipedia data in the "high-quality" reference texts used to develop the filter. 

    \textbf{AAL Sources.}
    To better understand how automated filters' predictions vary across AAL sources, \autoref{fig:filter-res} (bottom) presents the same average normalized model predictions on AAL texts from dialogues (\textit{CORAAL}), song lyrics (\textit{Hip Hop}), and social media (\textit{AAL Tweets}). As CORAAL contains natural speech and is unlikely to contain the lexical features exhibited by hip hop lyrics and social media, \textbf{most filters (11 of \numfilter{}) expectedly are more likely to conserve dialogue transcripts over the other sources. Most filters (12 of \numfilter{}) are also more likely to conserve AAL tweets over hip hop lyrics}. While filters appear to elevate natural speech, texts like transcripts in CORAAL are not widely available online suggesting that the sources of pretraining data may be the more influential factor determining AAL's representation in corpora (see \autoref{app:aal-urls} for further discussion).

\section{Discussion and Conclusion}

    In this work, we examine the quantity and quality of AAL representation in pretraining corpora as well as the impact of recent automated quality filters on such representation. 
    We first show that AAL %
    is widely underrepresented in modern corpora, and that many of those AAL documents are shared among corpora. We additionally highlight the ways in which AAL documents may be misrepresentative in these corpora, including analyses of highly performative language as well as documents that are judged to be written by non-AAL speakers, stereotypical, or inappropriate for LLMs' to generate. Finally, we show how automated filters used in recent pretraining corpora impact the quantity and quality of AAL representation.

    Our findings have several implications. The lack of AAL in pretraining corpora may underlie LLMs' difficulties in understanding AAL (e.g., \citet{groenwold-gen,deas-aal}), particularly for specific features \cite{kleiner-camouflage,grieser-deficiencies}. Beyond its underrepresentation, the misrepresentation of AAL in pretraining corpora may reinforce the discriminatory language model behaviors that have been identified in prior work \cite{fleisig-chatgpt,hofmann-covert} and restrict the benefits of LLMs for AAL speakers \cite{brewer-equitable}. Similar language technologies' inability to reliably interpret AAL has also been shown to create barriers to use of the technology \cite{cunningham-understanding,harrington-health}. Documenting the representation of AAL in pretraining corpora enables better understanding of why LLMs and other technologies pose such risks and offers insights into methods of mitigating them.

    While we find that automated filters are more likely to remove AAL than WME, they are also more likely to conserve naturalistic speech (CORAAL) over tweets and hip hop lyrics. Considering we highlight through RQ2 that highly performative AAL is prevalent in the corpora, this also suggests that there is a lack of naturalistic AAL in the sources leveraged for pretraining corpora. 
    Incorporating understanding of text sources in LLM development and carefully curating these sources through meaningful inclusion of AAL speakers as stakeholders in design processes \cite{friedman-participatory, suresh-foundation} is necessary to faithfully represent naturalistic AAL use. 
    We recommend that in developing LLMs designed for socially impactful domains where AAL speakers already face disparate impacts (e.g., medical \cite{beach-justice} or legal \cite{jones-testifying} settings), pretraining data curators ensure that the sources and filtering processes produce representative AAL data. 
    Finally, language understanding benchmarks that drive LLM development are unlikely to contain AAL, potentially leading to the similar underrepresentation of AAL found in pretraining corpora. The aforementioned considerations may also accordingly be extended to the benchmarks used to evaluate the effectiveness of design decisions throughout the LLM development cycle, incentivizing understanding of sociolinguistic variation as well as commonsense understanding and other capabilities.

\section{Limitations}

    We outline limitations to consider alongside our presented results. First, there are a wide variety of pretraining corpora and filtering strategies in use for LLM development, and the corpora we evaluate are not exhaustive. Notably, our work is enabled by the efforts to open-source language model development, but many corpora, such as those used to pretrain GPT-4o \cite{gpt4}, Llama-3 \cite{grattafiori2024llama3herdmodels}, and similarly large, closed-source models are unavailable for analysis. We include multiple corpora in our experiments, particularly those that attempt to replicate commercial corpora to better ensure that the findings are broadly applicable to current and future models. Additionally, given that online texts are the primary source of data for pertaining, we focus experiments on corpora composed of Common Crawl and internet sources.

    Furthermore, we collect human judgments of AAL text quality, though we acknowledge that our recruited annotators may not be fully representative of the perspectives and opinions of the broader AAL-speaking community. We recruit annotators that have experience in linguistics, NLP, and AAL as they are equipped to identify AAL features and consider the implications of using specific texts in LLM development. We hope that future work will consult diverse and representative annotators in both analyzing and curating data to better inform pretraining data collection.

    Finally, we follow prior work in using the demographic alignment classifier from \citet{blodgett-variation} to extract AAL from pretraining data for analyses. The classifier, however, is fitted on Twitter data which may not generalize to other language contexts. We broaden our analysis to consider any texts with higher than 0.3 probability of containing AAL to mitigate the impacts of the potential domain difference. 
    Similarly, the CGEdit classifier was not initially developed for internet texts. However, analyses using CGEdit focus on morphosyntactic constructions which are similarly portrayed online and in speech unlike features linked to orthography (e.g., \citealt{eisenstein-phon}).

\section{Ethics Statement}

\paragraph{Author Positionality.}
As a significant portion of this work involves qualitative analyses of texts and estimates, we acknowledge that the authors' backgrounds and perspectives may impact interpretation of some findings. In this work, we attempt to calculate and report quantitative statistics to complement qualitative discussion of the contents of pretraining corpora. As we strictly analyze open-source pretraining corpora, we encourage readers to examine texts within each corpus as well as form their own interpretations from the presented results. 

\paragraph{AAL in Pretraining and Benchmark Data.}
Our analyses show that AAL is underrepresented in all pretraining corpora studied, and that many documents are misrepresentative of naturalistic AAL. While increasing the representation of AAL in these corpora may improve downstream understanding in LLMs, at the same time, increased representation can pose risks to AAL speakers. As discussed by \citet{patton-casm}, consideration of the context of data and impact of technologies on vulnerable populations is necessary for research involving these communities. Particularly if data is collected without the consent and inclusion of AAL-speaking communities, increased representation in corpora may increase privacy risks \cite{brown-privacy} or perpetuate misrepresentative notions of AAL by means of LLMs.

\paragraph{AAL-use by LLMs.}
Our findings contribute to the ongoing study of AAL biases in both language understanding and generation systems. Prior work highlights the barriers that language technologies' lack of AAL understanding poses to African American users \cite{cunningham-understanding}, while others detail how human prejudices can lead to, for example, disproportionately labeling AAL as toxic \cite{sap-annot}. With respect to language \textit{generation}, recent work has found that African American users prefer and trust current chatbots using WME rather than AAL in generations \cite{finch-voice, basoah-anthropomorphic}. These findings highlight the conflict between language technologies that are able to properly understand AAL but not necessarily to generate AAL. As such, while our findings document factors in pretraining data that may lead to misunderstanding of AAL in LLMs, increasing representation alone does not absolve language model creators of their obligation to take in the culturally relevant context.

To conclude, no LLM depiction of AAL should misrepresent or caricature AAL use. In contexts where LLM use serves the social benefit of AAL speakers, measures should be taken to ensure that depictions of AAL do not trivialize the rich linguistic and cultural background surrounding them.

\section{Acknowledgments}
This work was supported in part by a Provost Diversity Fellow mini-grant, a grant from the Knight First Ammendment Institute at Columbia University, and grant IIS-2106666 from the National Science Foundation. 
The first author is supported by the National Science Foundation Graduate Research Fellowship DGE-2036197, the Columbia University Provost Diversity Fellowship, the Columbia School of Engineering and Applied Sciences Presidential Fellowship. Any opinion, findings, and conclusions or recommendations expressed in this material are those of the authors and do not necessarily reflect the views of the National Science Foundation or Knight First Amendment Institute. We thank the anonymous reviewers and the following people for providing valuable feedback on an earlier draft: Elsbeth Turcan, Matthew Toles, and Narutatsu Ri.

\bibliography{custom}

\begin{thebibliography}{88}
\providecommand{\natexlab}[1]{#1}

\bibitem[{Adilazuarda et~al.(2024)Adilazuarda, Mukherjee, Lavania, Singh, Aji, O{'}Neill, Modi, and Choudhury}]{adilazuarda-culture}
Muhammad~Farid Adilazuarda, Sagnik Mukherjee, Pradhyumna Lavania, Siddhant~Shivdutt Singh, Alham~Fikri Aji, Jacki O{'}Neill, Ashutosh Modi, and Monojit Choudhury. 2024.
\newblock \href {https://doi.org/10.18653/v1/2024.emnlp-main.882} {Towards measuring and modeling {``}culture{''} in {LLM}s: A survey}.
\newblock In \emph{Proceedings of the 2024 Conference on Empirical Methods in Natural Language Processing}, pages 15763--15784, Miami, Florida, USA. Association for Computational Linguistics.

\bibitem[{Alim(2006)}]{alim-roc}
H~Samy Alim. 2006.
\newblock \emph{Roc the mic right: The language of hip hop culture}.
\newblock Routledge.

\bibitem[{Baker-Bell(2020)}]{baker-bell-aal}
April Baker-Bell. 2020.
\newblock \emph{Linguistic justice: Black language, literacy, identity, and pedagogy}.
\newblock Routledge.

\bibitem[{Basoah et~al.(2025)Basoah, Chechelnitsky, Long, Reinecke, Zerva, Zhou, D{\'\i}az, and Sap}]{basoah-anthropomorphic}
Jeffrey Basoah, Daniel Chechelnitsky, Tao Long, Katharina Reinecke, Chrysoula Zerva, Kaitlyn Zhou, Mark D{\'\i}az, and Maarten Sap. 2025.
\newblock Not like us, hunty: Measuring perceptions and behavioral effects of minoritized anthropomorphic cues in llms.
\newblock \emph{arXiv preprint arXiv:2505.05660}.

\bibitem[{Beach et~al.(2021)Beach, Saha, Park, Taylor, Drew, Plank, Cooper, and Chee}]{beach-justice}
Mary~Catherine Beach, Somnath Saha, Jenny Park, Janiece Taylor, Paul Drew, Eve Plank, Lisa~A. Cooper, and Brant Chee. 2021.
\newblock \href {https://doi.org/10.1007/s11606-021-06682-z} {Testimonial injustice: Linguistic bias in the medical records of black patients and women}.
\newblock \emph{Journal of General Internal Medicine}, 36(6):1708–1714.

\bibitem[{Blodgett et~al.(2020)Blodgett, Barocas, Daum{\'e}~III, and Wallach}]{blodgett-harm}
Su~Lin Blodgett, Solon Barocas, Hal Daum{\'e}~III, and Hanna Wallach. 2020.
\newblock \href {https://doi.org/10.18653/v1/2020.acl-main.485} {Language (technology) is power: A critical survey of {``}bias{''} in {NLP}}.
\newblock In \emph{Proceedings of the 58th Annual Meeting of the Association for Computational Linguistics}, pages 5454--5476, Online. Association for Computational Linguistics.

\bibitem[{Blodgett et~al.(2016)Blodgett, Green, and O{'}Connor}]{blodgett-variation}
Su~Lin Blodgett, Lisa Green, and Brendan O{'}Connor. 2016.
\newblock \href {https://doi.org/10.18653/v1/D16-1120} {Demographic dialectal variation in social media: A case study of {A}frican-{A}merican {E}nglish}.
\newblock In \emph{Proceedings of the 2016 Conference on Empirical Methods in Natural Language Processing}, pages 1119--1130, Austin, Texas. Association for Computational Linguistics.

\bibitem[{Blodgett and O'Connor(2017)}]{blodgett-racial}
Su~Lin Blodgett and Brendan O'Connor. 2017.
\newblock Racial disparity in natural language processing: A case study of social media african-american english.
\newblock \emph{arXiv preprint arXiv:1707.00061}.

\bibitem[{Brewer et~al.(2023)Brewer, Harrington, and Heldreth}]{brewer-equitable}
Robin~N. Brewer, Christina Harrington, and Courtney Heldreth. 2023.
\newblock \href {https://doi.org/10.1145/3593013.3594005} {Envisioning equitable speech technologies for black older adults}.
\newblock In \emph{Proceedings of the 2023 ACM Conference on Fairness, Accountability, and Transparency}, FAccT '23, page 379–388, New York, NY, USA. Association for Computing Machinery.

\bibitem[{Brown et~al.(2022)Brown, Lee, Mireshghallah, Shokri, and Tram\`{e}r}]{brown-privacy}
Hannah Brown, Katherine Lee, Fatemehsadat Mireshghallah, Reza Shokri, and Florian Tram\`{e}r. 2022.
\newblock \href {https://doi.org/10.1145/3531146.3534642} {What does it mean for a language model to preserve privacy?}
\newblock In \emph{Proceedings of the 2022 ACM Conference on Fairness, Accountability, and Transparency}, FAccT '22, page 2280–2292, New York, NY, USA. Association for Computing Machinery.

\bibitem[{Brown et~al.(2020)Brown, Mann, Ryder, Subbiah, Kaplan, Dhariwal, Neelakantan, Shyam, Sastry, Askell, Agarwal, Herbert-Voss, Krueger, Henighan, Child, Ramesh, Ziegler, Wu, Winter, Hesse, Chen, Sigler, Litwin, Gray, Chess, Clark, Berner, McCandlish, Radford, Sutskever, and Amodei}]{brown-gpt-3}
Tom Brown, Benjamin Mann, Nick Ryder, Melanie Subbiah, Jared~D Kaplan, Prafulla Dhariwal, Arvind Neelakantan, Pranav Shyam, Girish Sastry, Amanda Askell, Sandhini Agarwal, Ariel Herbert-Voss, Gretchen Krueger, Tom Henighan, Rewon Child, Aditya Ramesh, Daniel Ziegler, Jeffrey Wu, Clemens Winter, Chris Hesse, Mark Chen, Eric Sigler, Mateusz Litwin, Scott Gray, Benjamin Chess, Jack Clark, Christopher Berner, Sam McCandlish, Alec Radford, Ilya Sutskever, and Dario Amodei. 2020.
\newblock \href {https://proceedings.neurips.cc/paper_files/paper/2020/file/1457c0d6bfcb4967418bfb8ac142f64a-Paper.pdf} {Language models are few-shot learners}.
\newblock In \emph{Advances in Neural Information Processing Systems}, volume~33, pages 1877--1901. Curran Associates, Inc.

\bibitem[{Campbell-Kibler(2008)}]{campbell-judge}
Kathryn Campbell-Kibler. 2008.
\newblock \href {https://doi.org/10.1017/S0047404508080974} {I’ll be the judge of that: Diversity in social perceptions of (ing)}.
\newblock \emph{Language in Society}, 37(5):637–659.

\bibitem[{Cheng et~al.(2022)Cheng, Mosallanezhad, Silva, Hall, and Liu}]{cheng-mitigate}
Lu~Cheng, Ahmadreza Mosallanezhad, Yasin~N. Silva, Deborah~L. Hall, and Huan Liu. 2022.
\newblock \href {https://doi.org/10.1145/3477495.3531945} {Bias mitigation for toxicity detection via sequential decisions}.
\newblock In \emph{Proceedings of the 45th International ACM SIGIR Conference on Research and Development in Information Retrieval}, SIGIR '22, page 1750–1760, New York, NY, USA. Association for Computing Machinery.

\bibitem[{Chowdhery et~al.(2022)Chowdhery, Narang, Devlin, Bosma, Mishra, Roberts, Barham, Chung, Sutton, Gehrmann, Schuh, Shi, Tsvyashchenko, Maynez, Rao, Barnes, Tay, Shazeer, Prabhakaran, Reif, Du, Hutchinson, Pope, Bradbury, Austin, Isard, Gur-Ari, Yin, Duke, Levskaya, Ghemawat, Dev, Michalewski, Garcia, Misra, Robinson, Fedus, Zhou, Ippolito, Luan, Lim, Zoph, Spiridonov, Sepassi, Dohan, Agrawal, Omernick, Dai, Pillai, Pellat, Lewkowycz, Moreira, Child, Polozov, Lee, Zhou, Wang, Saeta, Diaz, Firat, Catasta, Wei, Meier-Hellstern, Eck, Dean, Petrov, and Fiedel}]{chowdhery-palm}
Aakanksha Chowdhery, Sharan Narang, Jacob Devlin, Maarten Bosma, Gaurav Mishra, Adam Roberts, Paul Barham, Hyung~Won Chung, Charles Sutton, Sebastian Gehrmann, Parker Schuh, Kensen Shi, Sasha Tsvyashchenko, Joshua Maynez, Abhishek Rao, Parker Barnes, Yi~Tay, Noam Shazeer, Vinodkumar Prabhakaran, Emily Reif, Nan Du, Ben Hutchinson, Reiner Pope, James Bradbury, Jacob Austin, Michael Isard, Guy Gur-Ari, Pengcheng Yin, Toju Duke, Anselm Levskaya, Sanjay Ghemawat, Sunipa Dev, Henryk Michalewski, Xavier Garcia, Vedant Misra, Kevin Robinson, Liam Fedus, Denny Zhou, Daphne Ippolito, David Luan, Hyeontaek Lim, Barret Zoph, Alexander Spiridonov, Ryan Sepassi, David Dohan, Shivani Agrawal, Mark Omernick, Andrew~M. Dai, Thanumalayan~Sankaranarayana Pillai, Marie Pellat, Aitor Lewkowycz, Erica Moreira, Rewon Child, Oleksandr Polozov, Katherine Lee, Zongwei Zhou, Xuezhi Wang, Brennan Saeta, Mark Diaz, Orhan Firat, Michele Catasta, Jason Wei, Kathy Meier-Hellstern, Douglas Eck, Jeff Dean, Slav Petrov, and Noah Fiedel. 2022.
\newblock \href {https://arxiv.org/abs/2204.02311} {Palm: Scaling language modeling with pathways}.
\newblock \emph{Preprint}, arXiv:2204.02311.

\bibitem[{Clay(2003)}]{clay-hip-hop}
Andreana Clay. 2003.
\newblock \href {https://doi.org/10.1177/0002764203046010005} {Keepin' it real: Black youth, hip-hop culture, and black identity}.
\newblock \emph{American Behavioral Scientist}, 46(10):1346--1358.

\bibitem[{Computer(2023)}]{redpajama}
Together Computer. 2023.
\newblock \href {https://github.com/togethercomputer/RedPajama-Data} {Redpajama: An open source recipe to reproduce llama training dataset}.

\bibitem[{Corradini(2024)}]{corradini-approp}
Enrico Corradini. 2024.
\newblock \href {https://doi.org/10.1016/j.ipm.2024.103662} {Deconstructing cultural appropriation in online communities: A multilayer network analysis approach}.
\newblock \emph{Information Processing \& Management}, 61(3):103662.

\bibitem[{Cunningham et~al.(2024)Cunningham, Blodgett, Madaio, Daum{\'e}~Iii, Harrington, and Wallach}]{cunningham-understanding}
Jay Cunningham, Su~Lin Blodgett, Michael Madaio, Hal Daum{\'e}~Iii, Christina Harrington, and Hanna Wallach. 2024.
\newblock \href {https://doi.org/10.18653/v1/2024.findings-acl.761} {Understanding the impacts of language technologies{'} performance disparities on {A}frican {A}merican language speakers}.
\newblock In \emph{Findings of the Association for Computational Linguistics ACL 2024}, pages 12826--12833, Bangkok, Thailand and virtual meeting. Association for Computational Linguistics.

\bibitem[{Cunningham(2014)}]{cunningham-daal}
Jennifer~M. Cunningham. 2014.
\newblock \href {https://doi.org/10.1177/0741088314547188} {Features of digital african american language in a social network site}.
\newblock \emph{Written Communication}, 31(4):404--433.

\bibitem[{Dacon et~al.(2022)Dacon, Liu, and Tang}]{dacon-nli}
Jamell Dacon, Haochen Liu, and Jiliang Tang. 2022.
\newblock \href {https://aclanthology.org/2022.coling-1.124} {Evaluating and mitigating inherent linguistic bias of {A}frican {A}merican {E}nglish through inference}.
\newblock In \emph{Proceedings of the 29th International Conference on Computational Linguistics}, pages 1442--1454, Gyeongju, Republic of Korea. International Committee on Computational Linguistics.

\bibitem[{Davidson et~al.(2019)Davidson, Bhattacharya, and Weber}]{davidson-datasets}
Thomas Davidson, Debasmita Bhattacharya, and Ingmar Weber. 2019.
\newblock \href {https://doi.org/10.18653/v1/W19-3504} {Racial bias in hate speech and abusive language detection datasets}.
\newblock In \emph{Proceedings of the Third Workshop on Abusive Language Online}, pages 25--35, Florence, Italy. Association for Computational Linguistics.

\bibitem[{Davidson-Pilon(2015)}]{tdigest}
Cameron Davidson-Pilon. 2015.
\newblock t-digest.
\newblock \url{https://github.com/CamDavidsonPilon/tdigest}.

\bibitem[{Deas et~al.(2024)Deas, Grieser, Hou, Kleiner, Martin, Nandanampati, Patton, and McKeown}]{deas-phon}
Nicholas Deas, Jessi Grieser, Xinmeng Hou, Shana Kleiner, Tajh Martin, Sreya Nandanampati, Desmond Patton, and Kathleen McKeown. 2024.
\newblock Phonate: Impact of type-written phonological features of african american language on generative language model- ing tasks.
\newblock In \emph{Proceedings of the First Conference on Language Modeling}.

\bibitem[{Deas et~al.(2023)Deas, Grieser, Kleiner, Patton, Turcan, and McKeown}]{deas-aal}
Nicholas Deas, Jessica Grieser, Shana Kleiner, Desmond Patton, Elsbeth Turcan, and Kathleen McKeown. 2023.
\newblock \href {https://doi.org/10.18653/v1/2023.emnlp-main.421} {Evaluation of {A}frican {A}merican language bias in natural language generation}.
\newblock In \emph{Proceedings of the 2023 Conference on Empirical Methods in Natural Language Processing}, pages 6805--6824, Singapore. Association for Computational Linguistics.

\bibitem[{Dodge et~al.(2021)Dodge, Sap, Marasovi{\'c}, Agnew, Ilharco, Groeneveld, Mitchell, and Gardner}]{dodge-c4}
Jesse Dodge, Maarten Sap, Ana Marasovi{\'c}, William Agnew, Gabriel Ilharco, Dirk Groeneveld, Margaret Mitchell, and Matt Gardner. 2021.
\newblock \href {https://doi.org/10.18653/v1/2021.emnlp-main.98} {Documenting large webtext corpora: A case study on the colossal clean crawled corpus}.
\newblock In \emph{Proceedings of the 2021 Conference on Empirical Methods in Natural Language Processing}, pages 1286--1305, Online and Punta Cana, Dominican Republic. Association for Computational Linguistics.

\bibitem[{Eisenstein(2013)}]{eisenstein-phon}
Jacob Eisenstein. 2013.
\newblock \href {https://aclanthology.org/W13-1102} {Phonological factors in social media writing}.
\newblock In \emph{Proceedings of the Workshop on Language Analysis in Social Media}, pages 11--19, Atlanta, Georgia. Association for Computational Linguistics.

\bibitem[{Farrington and Kendall(2021)}]{farrington-coraal}
Charlie Farrington and Tyler Kendall. 2021.
\newblock \href {https://doi.org/10.7264/1AD5-6T35} {The corpus of regional african american language}.

\bibitem[{Feng et~al.(2023)Feng, Park, Liu, and Tsvetkov}]{feng-political}
Shangbin Feng, Chan~Young Park, Yuhan Liu, and Yulia Tsvetkov. 2023.
\newblock \href {https://doi.org/10.18653/v1/2023.acl-long.656} {From pretraining data to language models to downstream tasks: Tracking the trails of political biases leading to unfair {NLP} models}.
\newblock In \emph{Proceedings of the 61st Annual Meeting of the Association for Computational Linguistics (Volume 1: Long Papers)}, pages 11737--11762, Toronto, Canada. Association for Computational Linguistics.

\bibitem[{Finch et~al.(2025)Finch, Paek, Kwon, Choi, Wells, Chandler, and Choi}]{finch-voice}
Sarah~E. Finch, Ellie~S. Paek, Sejung Kwon, Ikseon Choi, Jessica Wells, Rasheeta Chandler, and Jinho~D. Choi. 2025.
\newblock \href {https://arxiv.org/abs/2501.03441} {Finding a voice: Evaluating african american dialect generation for chatbot technology}.
\newblock \emph{Preprint}, arXiv:2501.03441.

\bibitem[{Fleisig et~al.(2024)Fleisig, Smith, Bossi, Rustagi, Yin, and Klein}]{fleisig-chatgpt}
Eve Fleisig, Genevieve Smith, Madeline Bossi, Ishita Rustagi, Xavier Yin, and Dan Klein. 2024.
\newblock \href {https://arxiv.org/abs/2406.08818} {Linguistic bias in chatgpt: Language models reinforce dialect discrimination}.
\newblock \emph{Preprint}, arXiv:2406.08818.

\bibitem[{Friedman(1996)}]{friedman-participatory}
Batya Friedman. 1996.
\newblock \href {https://doi.org/10.1145/242485.242493} {Value-sensitive design}.
\newblock \emph{Interactions}, 3(6):16–23.

\bibitem[{Gao et~al.(2020)Gao, Biderman, Black, Golding, Hoppe, Foster, Phang, He, Thite, Nabeshima, Presser, and Leahy}]{thepile}
Leo Gao, Stella Biderman, Sid Black, Laurence Golding, Travis Hoppe, Charles Foster, Jason Phang, Horace He, Anish Thite, Noa Nabeshima, Shawn Presser, and Connor Leahy. 2020.
\newblock \href {https://arxiv.org/abs/2101.00027} {The pile: An 800gb dataset of diverse text for language modeling}.
\newblock \emph{Preprint}, arXiv:2101.00027.

\bibitem[{Gokaslan et~al.(2019)Gokaslan, Cohen, Pavlick, and Tellex}]{gokaslan-openwebtext}
Aaron Gokaslan, Vanya Cohen, Ellie Pavlick, and Stefanie Tellex. 2019.
\newblock Openwebtext corpus.
\newblock \url{http://Skylion007.github.io/OpenWebTextCorpus}.

\bibitem[{Grattafiori et~al.(2024)Grattafiori, Dubey, Jauhri, Pandey, Kadian, Al-Dahle, Letman, Mathur, Schelten, Vaughan, Yang, Fan, Goyal, Hartshorn, Yang, Mitra, Sravankumar, Korenev, Hinsvark, Rao, Zhang, Rodriguez, Gregerson, Spataru, Roziere, Biron, Tang, Chern, Caucheteux, Nayak, Bi, Marra, McConnell, Keller, Touret, Wu, Wong, Ferrer, Nikolaidis, Allonsius, Song, Pintz, Livshits, Wyatt, Esiobu, Choudhary, Mahajan, Garcia-Olano, Perino, Hupkes, Lakomkin, AlBadawy, Lobanova, Dinan, Smith, Radenovic, Guzmán, Zhang, Synnaeve, Lee, Anderson, Thattai, Nail, Mialon, Pang, Cucurell, Nguyen, Korevaar, Xu, Touvron, Zarov, Ibarra, Kloumann, Misra, Evtimov, Zhang, Copet, Lee, Geffert, Vranes, Park, Mahadeokar, Shah, van~der Linde, Billock, Hong, Lee, Fu, Chi, Huang, Liu, Wang, Yu, Bitton, Spisak, Park, Rocca, Johnstun, Saxe, Jia, Alwala, Prasad, Upasani, Plawiak, Li, Heafield, Stone, El-Arini, Iyer, Malik, Chiu, Bhalla, Lakhotia, Rantala-Yeary, van~der Maaten, Chen, Tan, Jenkins, Martin, Madaan, Malo, Blecher,
  Landzaat, de~Oliveira, Muzzi, Pasupuleti, Singh, Paluri, Kardas, Tsimpoukelli, Oldham, Rita, Pavlova, Kambadur, Lewis, Si, Singh, Hassan, Goyal, Torabi, Bashlykov, Bogoychev, Chatterji, Zhang, Duchenne, Çelebi, Alrassy, Zhang, Li, Vasic, Weng, Bhargava, Dubal, Krishnan, Koura, Xu, He, Dong, Srinivasan, Ganapathy, Calderer, Cabral, Stojnic, Raileanu, Maheswari, Girdhar, Patel, Sauvestre, Polidoro, Sumbaly, Taylor, Silva, Hou, Wang, Hosseini, Chennabasappa, Singh, Bell, Kim, Edunov, Nie, Narang, Raparthy, Shen, Wan, Bhosale, Zhang, Vandenhende, Batra, Whitman, Sootla, Collot, Gururangan, Borodinsky, Herman, Fowler, Sheasha, Georgiou, Scialom, Speckbacher, Mihaylov, Xiao, Karn, Goswami, Gupta, Ramanathan, Kerkez, Gonguet, Do, Vogeti, Albiero, Petrovic, Chu, Xiong, Fu, Meers, Martinet, Wang, Wang, Tan, Xia, Xie, Jia, Wang, Goldschlag, Gaur, Babaei, Wen, Song, Zhang, Li, Mao, Coudert, Yan, Chen, Papakipos, Singh, Srivastava, Jain, Kelsey, Shajnfeld, Gangidi, Victoria, Goldstand, Menon, Sharma, Boesenberg,
  Baevski, Feinstein, Kallet, Sangani, Teo, Yunus, Lupu, Alvarado, Caples, Gu, Ho, Poulton, Ryan, Ramchandani, Dong, Franco, Goyal, Saraf, Chowdhury, Gabriel, Bharambe, Eisenman, Yazdan, James, Maurer, Leonhardi, Huang, Loyd, Paola, Paranjape, Liu, Wu, Ni, Hancock, Wasti, Spence, Stojkovic, Gamido, Montalvo, Parker, Burton, Mejia, Liu, Wang, Kim, Zhou, Hu, Chu, Cai, Tindal, Feichtenhofer, Gao, Civin, Beaty, Kreymer, Li, Adkins, Xu, Testuggine, David, Parikh, Liskovich, Foss, Wang, Le, Holland, Dowling, Jamil, Montgomery, Presani, Hahn, Wood, Le, Brinkman, Arcaute, Dunbar, Smothers, Sun, Kreuk, Tian, Kokkinos, Ozgenel, Caggioni, Kanayet, Seide, Florez, Schwarz, Badeer, Swee, Halpern, Herman, Sizov, Guangyi, Zhang, Lakshminarayanan, Inan, Shojanazeri, Zou, Wang, Zha, Habeeb, Rudolph, Suk, Aspegren, Goldman, Zhan, Damlaj, Molybog, Tufanov, Leontiadis, Veliche, Gat, Weissman, Geboski, Kohli, Lam, Asher, Gaya, Marcus, Tang, Chan, Zhen, Reizenstein, Teboul, Zhong, Jin, Yang, Cummings, Carvill, Shepard, McPhie,
  Torres, Ginsburg, Wang, Wu, U, Saxena, Khandelwal, Zand, Matosich, Veeraraghavan, Michelena, Li, Jagadeesh, Huang, Chawla, Huang, Chen, Garg, A, Silva, Bell, Zhang, Guo, Yu, Moshkovich, Wehrstedt, Khabsa, Avalani, Bhatt, Mankus, Hasson, Lennie, Reso, Groshev, Naumov, Lathi, Keneally, Liu, Seltzer, Valko, Restrepo, Patel, Vyatskov, Samvelyan, Clark, Macey, Wang, Hermoso, Metanat, Rastegari, Bansal, Santhanam, Parks, White, Bawa, Singhal, Egebo, Usunier, Mehta, Laptev, Dong, Cheng, Chernoguz, Hart, Salpekar, Kalinli, Kent, Parekh, Saab, Balaji, Rittner, Bontrager, Roux, Dollar, Zvyagina, Ratanchandani, Yuvraj, Liang, Alao, Rodriguez, Ayub, Murthy, Nayani, Mitra, Parthasarathy, Li, Hogan, Battey, Wang, Howes, Rinott, Mehta, Siby, Bondu, Datta, Chugh, Hunt, Dhillon, Sidorov, Pan, Mahajan, Verma, Yamamoto, Ramaswamy, Lindsay, Lindsay, Feng, Lin, Zha, Patil, Shankar, Zhang, Zhang, Wang, Agarwal, Sajuyigbe, Chintala, Max, Chen, Kehoe, Satterfield, Govindaprasad, Gupta, Deng, Cho, Virk, Subramanian, Choudhury,
  Goldman, Remez, Glaser, Best, Koehler, Robinson, Li, Zhang, Matthews, Chou, Shaked, Vontimitta, Ajayi, Montanez, Mohan, Kumar, Mangla, Ionescu, Poenaru, Mihailescu, Ivanov, Li, Wang, Jiang, Bouaziz, Constable, Tang, Wu, Wang, Wu, Gao, Kleinman, Chen, Hu, Jia, Qi, Li, Zhang, Zhang, Adi, Nam, Yu, Wang, Zhao, Hao, Qian, Li, He, Rait, DeVito, Rosnbrick, Wen, Yang, Zhao, and Ma}]{grattafiori2024llama3herdmodels}
Aaron Grattafiori, Abhimanyu Dubey, Abhinav Jauhri, Abhinav Pandey, Abhishek Kadian, Ahmad Al-Dahle, Aiesha Letman, Akhil Mathur, Alan Schelten, Alex Vaughan, Amy Yang, Angela Fan, Anirudh Goyal, Anthony Hartshorn, Aobo Yang, Archi Mitra, Archie Sravankumar, Artem Korenev, Arthur Hinsvark, Arun Rao, Aston Zhang, Aurelien Rodriguez, Austen Gregerson, Ava Spataru, Baptiste Roziere, Bethany Biron, Binh Tang, Bobbie Chern, Charlotte Caucheteux, Chaya Nayak, Chloe Bi, Chris Marra, Chris McConnell, Christian Keller, Christophe Touret, Chunyang Wu, Corinne Wong, Cristian~Canton Ferrer, Cyrus Nikolaidis, Damien Allonsius, Daniel Song, Danielle Pintz, Danny Livshits, Danny Wyatt, David Esiobu, Dhruv Choudhary, Dhruv Mahajan, Diego Garcia-Olano, Diego Perino, Dieuwke Hupkes, Egor Lakomkin, Ehab AlBadawy, Elina Lobanova, Emily Dinan, Eric~Michael Smith, Filip Radenovic, Francisco Guzmán, Frank Zhang, Gabriel Synnaeve, Gabrielle Lee, Georgia~Lewis Anderson, Govind Thattai, Graeme Nail, Gregoire Mialon, Guan Pang,
  Guillem Cucurell, Hailey Nguyen, Hannah Korevaar, Hu~Xu, Hugo Touvron, Iliyan Zarov, Imanol~Arrieta Ibarra, Isabel Kloumann, Ishan Misra, Ivan Evtimov, Jack Zhang, Jade Copet, Jaewon Lee, Jan Geffert, Jana Vranes, Jason Park, Jay Mahadeokar, Jeet Shah, Jelmer van~der Linde, Jennifer Billock, Jenny Hong, Jenya Lee, Jeremy Fu, Jianfeng Chi, Jianyu Huang, Jiawen Liu, Jie Wang, Jiecao Yu, Joanna Bitton, Joe Spisak, Jongsoo Park, Joseph Rocca, Joshua Johnstun, Joshua Saxe, Junteng Jia, Kalyan~Vasuden Alwala, Karthik Prasad, Kartikeya Upasani, Kate Plawiak, Ke~Li, Kenneth Heafield, Kevin Stone, Khalid El-Arini, Krithika Iyer, Kshitiz Malik, Kuenley Chiu, Kunal Bhalla, Kushal Lakhotia, Lauren Rantala-Yeary, Laurens van~der Maaten, Lawrence Chen, Liang Tan, Liz Jenkins, Louis Martin, Lovish Madaan, Lubo Malo, Lukas Blecher, Lukas Landzaat, Luke de~Oliveira, Madeline Muzzi, Mahesh Pasupuleti, Mannat Singh, Manohar Paluri, Marcin Kardas, Maria Tsimpoukelli, Mathew Oldham, Mathieu Rita, Maya Pavlova, Melanie Kambadur,
  Mike Lewis, Min Si, Mitesh~Kumar Singh, Mona Hassan, Naman Goyal, Narjes Torabi, Nikolay Bashlykov, Nikolay Bogoychev, Niladri Chatterji, Ning Zhang, Olivier Duchenne, Onur Çelebi, Patrick Alrassy, Pengchuan Zhang, Pengwei Li, Petar Vasic, Peter Weng, Prajjwal Bhargava, Pratik Dubal, Praveen Krishnan, Punit~Singh Koura, Puxin Xu, Qing He, Qingxiao Dong, Ragavan Srinivasan, Raj Ganapathy, Ramon Calderer, Ricardo~Silveira Cabral, Robert Stojnic, Roberta Raileanu, Rohan Maheswari, Rohit Girdhar, Rohit Patel, Romain Sauvestre, Ronnie Polidoro, Roshan Sumbaly, Ross Taylor, Ruan Silva, Rui Hou, Rui Wang, Saghar Hosseini, Sahana Chennabasappa, Sanjay Singh, Sean Bell, Seohyun~Sonia Kim, Sergey Edunov, Shaoliang Nie, Sharan Narang, Sharath Raparthy, Sheng Shen, Shengye Wan, Shruti Bhosale, Shun Zhang, Simon Vandenhende, Soumya Batra, Spencer Whitman, Sten Sootla, Stephane Collot, Suchin Gururangan, Sydney Borodinsky, Tamar Herman, Tara Fowler, Tarek Sheasha, Thomas Georgiou, Thomas Scialom, Tobias Speckbacher,
  Todor Mihaylov, Tong Xiao, Ujjwal Karn, Vedanuj Goswami, Vibhor Gupta, Vignesh Ramanathan, Viktor Kerkez, Vincent Gonguet, Virginie Do, Vish Vogeti, Vítor Albiero, Vladan Petrovic, Weiwei Chu, Wenhan Xiong, Wenyin Fu, Whitney Meers, Xavier Martinet, Xiaodong Wang, Xiaofang Wang, Xiaoqing~Ellen Tan, Xide Xia, Xinfeng Xie, Xuchao Jia, Xuewei Wang, Yaelle Goldschlag, Yashesh Gaur, Yasmine Babaei, Yi~Wen, Yiwen Song, Yuchen Zhang, Yue Li, Yuning Mao, Zacharie~Delpierre Coudert, Zheng Yan, Zhengxing Chen, Zoe Papakipos, Aaditya Singh, Aayushi Srivastava, Abha Jain, Adam Kelsey, Adam Shajnfeld, Adithya Gangidi, Adolfo Victoria, Ahuva Goldstand, Ajay Menon, Ajay Sharma, Alex Boesenberg, Alexei Baevski, Allie Feinstein, Amanda Kallet, Amit Sangani, Amos Teo, Anam Yunus, Andrei Lupu, Andres Alvarado, Andrew Caples, Andrew Gu, Andrew Ho, Andrew Poulton, Andrew Ryan, Ankit Ramchandani, Annie Dong, Annie Franco, Anuj Goyal, Aparajita Saraf, Arkabandhu Chowdhury, Ashley Gabriel, Ashwin Bharambe, Assaf Eisenman, Azadeh
  Yazdan, Beau James, Ben Maurer, Benjamin Leonhardi, Bernie Huang, Beth Loyd, Beto~De Paola, Bhargavi Paranjape, Bing Liu, Bo~Wu, Boyu Ni, Braden Hancock, Bram Wasti, Brandon Spence, Brani Stojkovic, Brian Gamido, Britt Montalvo, Carl Parker, Carly Burton, Catalina Mejia, Ce~Liu, Changhan Wang, Changkyu Kim, Chao Zhou, Chester Hu, Ching-Hsiang Chu, Chris Cai, Chris Tindal, Christoph Feichtenhofer, Cynthia Gao, Damon Civin, Dana Beaty, Daniel Kreymer, Daniel Li, David Adkins, David Xu, Davide Testuggine, Delia David, Devi Parikh, Diana Liskovich, Didem Foss, Dingkang Wang, Duc Le, Dustin Holland, Edward Dowling, Eissa Jamil, Elaine Montgomery, Eleonora Presani, Emily Hahn, Emily Wood, Eric-Tuan Le, Erik Brinkman, Esteban Arcaute, Evan Dunbar, Evan Smothers, Fei Sun, Felix Kreuk, Feng Tian, Filippos Kokkinos, Firat Ozgenel, Francesco Caggioni, Frank Kanayet, Frank Seide, Gabriela~Medina Florez, Gabriella Schwarz, Gada Badeer, Georgia Swee, Gil Halpern, Grant Herman, Grigory Sizov, Guangyi, Zhang, Guna
  Lakshminarayanan, Hakan Inan, Hamid Shojanazeri, Han Zou, Hannah Wang, Hanwen Zha, Haroun Habeeb, Harrison Rudolph, Helen Suk, Henry Aspegren, Hunter Goldman, Hongyuan Zhan, Ibrahim Damlaj, Igor Molybog, Igor Tufanov, Ilias Leontiadis, Irina-Elena Veliche, Itai Gat, Jake Weissman, James Geboski, James Kohli, Janice Lam, Japhet Asher, Jean-Baptiste Gaya, Jeff Marcus, Jeff Tang, Jennifer Chan, Jenny Zhen, Jeremy Reizenstein, Jeremy Teboul, Jessica Zhong, Jian Jin, Jingyi Yang, Joe Cummings, Jon Carvill, Jon Shepard, Jonathan McPhie, Jonathan Torres, Josh Ginsburg, Junjie Wang, Kai Wu, Kam~Hou U, Karan Saxena, Kartikay Khandelwal, Katayoun Zand, Kathy Matosich, Kaushik Veeraraghavan, Kelly Michelena, Keqian Li, Kiran Jagadeesh, Kun Huang, Kunal Chawla, Kyle Huang, Lailin Chen, Lakshya Garg, Lavender A, Leandro Silva, Lee Bell, Lei Zhang, Liangpeng Guo, Licheng Yu, Liron Moshkovich, Luca Wehrstedt, Madian Khabsa, Manav Avalani, Manish Bhatt, Martynas Mankus, Matan Hasson, Matthew Lennie, Matthias Reso, Maxim
  Groshev, Maxim Naumov, Maya Lathi, Meghan Keneally, Miao Liu, Michael~L. Seltzer, Michal Valko, Michelle Restrepo, Mihir Patel, Mik Vyatskov, Mikayel Samvelyan, Mike Clark, Mike Macey, Mike Wang, Miquel~Jubert Hermoso, Mo~Metanat, Mohammad Rastegari, Munish Bansal, Nandhini Santhanam, Natascha Parks, Natasha White, Navyata Bawa, Nayan Singhal, Nick Egebo, Nicolas Usunier, Nikhil Mehta, Nikolay~Pavlovich Laptev, Ning Dong, Norman Cheng, Oleg Chernoguz, Olivia Hart, Omkar Salpekar, Ozlem Kalinli, Parkin Kent, Parth Parekh, Paul Saab, Pavan Balaji, Pedro Rittner, Philip Bontrager, Pierre Roux, Piotr Dollar, Polina Zvyagina, Prashant Ratanchandani, Pritish Yuvraj, Qian Liang, Rachad Alao, Rachel Rodriguez, Rafi Ayub, Raghotham Murthy, Raghu Nayani, Rahul Mitra, Rangaprabhu Parthasarathy, Raymond Li, Rebekkah Hogan, Robin Battey, Rocky Wang, Russ Howes, Ruty Rinott, Sachin Mehta, Sachin Siby, Sai~Jayesh Bondu, Samyak Datta, Sara Chugh, Sara Hunt, Sargun Dhillon, Sasha Sidorov, Satadru Pan, Saurabh Mahajan,
  Saurabh Verma, Seiji Yamamoto, Sharadh Ramaswamy, Shaun Lindsay, Shaun Lindsay, Sheng Feng, Shenghao Lin, Shengxin~Cindy Zha, Shishir Patil, Shiva Shankar, Shuqiang Zhang, Shuqiang Zhang, Sinong Wang, Sneha Agarwal, Soji Sajuyigbe, Soumith Chintala, Stephanie Max, Stephen Chen, Steve Kehoe, Steve Satterfield, Sudarshan Govindaprasad, Sumit Gupta, Summer Deng, Sungmin Cho, Sunny Virk, Suraj Subramanian, Sy~Choudhury, Sydney Goldman, Tal Remez, Tamar Glaser, Tamara Best, Thilo Koehler, Thomas Robinson, Tianhe Li, Tianjun Zhang, Tim Matthews, Timothy Chou, Tzook Shaked, Varun Vontimitta, Victoria Ajayi, Victoria Montanez, Vijai Mohan, Vinay~Satish Kumar, Vishal Mangla, Vlad Ionescu, Vlad Poenaru, Vlad~Tiberiu Mihailescu, Vladimir Ivanov, Wei Li, Wenchen Wang, Wenwen Jiang, Wes Bouaziz, Will Constable, Xiaocheng Tang, Xiaojian Wu, Xiaolan Wang, Xilun Wu, Xinbo Gao, Yaniv Kleinman, Yanjun Chen, Ye~Hu, Ye~Jia, Ye~Qi, Yenda Li, Yilin Zhang, Ying Zhang, Yossi Adi, Youngjin Nam, Yu, Wang, Yu~Zhao, Yuchen Hao, Yundi
  Qian, Yunlu Li, Yuzi He, Zach Rait, Zachary DeVito, Zef Rosnbrick, Zhaoduo Wen, Zhenyu Yang, Zhiwei Zhao, and Zhiyu Ma. 2024.
\newblock \href {https://arxiv.org/abs/2407.21783} {The llama 3 herd of models}.
\newblock \emph{Preprint}, arXiv:2407.21783.

\bibitem[{Green(2002)}]{green-aae}
Lisa~J Green. 2002.
\newblock \emph{African American English: a linguistic introduction}.
\newblock Cambridge University Press.

\bibitem[{Grieser(2022)}]{grieser-dc}
Jessica~A Grieser. 2022.
\newblock \emph{The Black side of the river: Race, language, and belonging in Washington, DC}.
\newblock Georgetown University Press.

\bibitem[{Grieser et~al.(2024)Grieser, Shepard, Deas, Kleiner, Patton, Turcan, and McKeown}]{grieser-deficiencies}
Jessica~A Grieser, James Shepard, Nicholas Deas, Shana Kleiner, Desmond Patton, Elsbeth Turcan, and Kathleen McKeown. 2024.
\newblock Exploring the deficiencies of large language models in use of aal: An interdisciplinary study.
\newblock \emph{Ann Arbor: University of Michigan, ms}.

\bibitem[{Groenwold et~al.(2020)Groenwold, Ou, Parekh, Honnavalli, Levy, Mirza, and Wang}]{groenwold-gen}
Sophie Groenwold, Lily Ou, Aesha Parekh, Samhita Honnavalli, Sharon Levy, Diba Mirza, and William~Yang Wang. 2020.
\newblock \href {https://doi.org/10.18653/v1/2020.emnlp-main.473} {Investigating {A}frican-{A}merican {V}ernacular {E}nglish in transformer-based text generation}.
\newblock In \emph{Proceedings of the 2020 Conference on Empirical Methods in Natural Language Processing (EMNLP)}, pages 5877--5883, Online. Association for Computational Linguistics.

\bibitem[{Gunasekar et~al.(2023)Gunasekar, Zhang, Aneja, Mendes, Del~Giorno, Gopi, Javaheripi, Kauffmann, de~Rosa, Saarikivi et~al.}]{gunasekar-textbooks}
Suriya Gunasekar, Yi~Zhang, Jyoti Aneja, Caio C{\'e}sar~Teodoro Mendes, Allie Del~Giorno, Sivakanth Gopi, Mojan Javaheripi, Piero Kauffmann, Gustavo de~Rosa, Olli Saarikivi, et~al. 2023.
\newblock Textbooks are all you need.
\newblock \emph{arXiv preprint arXiv:2306.11644}.

\bibitem[{Halevy et~al.(2021)Halevy, Harris, Bruckman, Yang, and Howard}]{halevy-mitigate}
Matan Halevy, Camille Harris, Amy Bruckman, Diyi Yang, and Ayanna Howard. 2021.
\newblock \href {https://doi.org/10.1145/3465416.3483299} {Mitigating racial biases in toxic language detection with an equity-based ensemble framework}.
\newblock In \emph{Proceedings of the 1st ACM Conference on Equity and Access in Algorithms, Mechanisms, and Optimization}, EAAMO '21, New York, NY, USA. Association for Computing Machinery.

\bibitem[{Harrington et~al.(2022)Harrington, Garg, Woodward, and Williams}]{harrington-health}
Christina~N. Harrington, Radhika Garg, Amanda Woodward, and Dimitri Williams. 2022.
\newblock \href {https://doi.org/10.1145/3491102.3501995} {“it’s kind of like code-switching”: Black older adults’ experiences with a voice assistant for health information seeking}.
\newblock In \emph{Proceedings of the 2022 CHI Conference on Human Factors in Computing Systems}, CHI '22, New York, NY, USA. Association for Computing Machinery.

\bibitem[{Harris et~al.(2022)Harris, Halevy, Howard, Bruckman, and Yang}]{harris-role}
Camille Harris, Matan Halevy, Ayanna Howard, Amy Bruckman, and Diyi Yang. 2022.
\newblock \href {https://doi.org/10.1145/3531146.3533144} {Exploring the role of grammar and word choice in bias toward african american english (aae) in hate speech classification}.
\newblock In \emph{Proceedings of the 2022 ACM Conference on Fairness, Accountability, and Transparency}, FAccT '22, page 789–798, New York, NY, USA. Association for Computing Machinery.

\bibitem[{Heafield(2011)}]{heafield-kenlm}
Kenneth Heafield. 2011.
\newblock \href {https://aclanthology.org/W11-2123/} {{K}en{LM}: Faster and smaller language model queries}.
\newblock In \emph{Proceedings of the Sixth Workshop on Statistical Machine Translation}, pages 187--197, Edinburgh, Scotland. Association for Computational Linguistics.

\bibitem[{Hendrycks et~al.(2021)Hendrycks, Burns, Basart, Zou, Mazeika, Song, and Steinhardt}]{hendrycks-mmlu}
Dan Hendrycks, Collin Burns, Steven Basart, Andy Zou, Mantas Mazeika, Dawn Song, and Jacob Steinhardt. 2021.
\newblock Measuring massive multitask language understanding.
\newblock \emph{Proceedings of the International Conference on Learning Representations (ICLR)}.

\bibitem[{Hofmann et~al.(2024)Hofmann, Kalluri, Jurafsky, and King}]{hofmann-covert}
Valentin Hofmann, Pratyusha~Ria Kalluri, Dan Jurafsky, and Sharese King. 2024.
\newblock \href {https://doi.org/10.1038/s41586-024-07856-5} {Ai generates covertly racist decisions about people based on their dialect}.
\newblock \emph{Nature}, 633(8028):147–154.

\bibitem[{Hovy and Prabhumoye(2021)}]{hovy-sources}
Dirk Hovy and Shrimai Prabhumoye. 2021.
\newblock \href {https://doi.org/10.1111/lnc3.12432} {Five sources of bias in natural language processing}.
\newblock \emph{Language and Linguistics Compass}, 15(8).

\bibitem[{Hunter(2007)}]{Hunter:2007}
J.~D. Hunter. 2007.
\newblock \href {https://doi.org/10.1109/MCSE.2007.55} {Matplotlib: A 2d graphics environment}.
\newblock \emph{Computing in Science \& Engineering}, 9(3):90--95.

\bibitem[{Ilbury(2020)}]{ilbury-sassy}
Christian Ilbury. 2020.
\newblock “sassy queens”: Stylistic orthographic variation in twitter and the enregisterment of aave.
\newblock \emph{Journal of sociolinguistics}, 24(2):245--264.

\bibitem[{Jones et~al.(2019)Jones, Kalbfeld, Hancock, and Clark}]{jones-testifying}
Taylor Jones, Jessica~Rose Kalbfeld, Ryan Hancock, and Robin Clark. 2019.
\newblock \href {https://doi.org/10.1353/lan.2019.0042} {Testifying while black: An experimental study of court reporter accuracy in transcription of african american english}.
\newblock \emph{Language}, 95(2):e216–e252.

\bibitem[{Joulin et~al.(2017)Joulin, Grave, Bojanowski, and Mikolov}]{joulin-fasttext}
Armand Joulin, Edouard Grave, Piotr Bojanowski, and Tomas Mikolov. 2017.
\newblock Bag of tricks for efficient text classification.
\newblock In \emph{Proceedings of the 15th Conference of the European Chapter of the Association for Computational Linguistics: Volume 2, Short Papers}, pages 427--431. Association for Computational Linguistics.

\bibitem[{Keswani and Celis(2021)}]{keswani-summ}
Vijay Keswani and L.~Elisa Celis. 2021.
\newblock \href {https://doi.org/10.1145/3442381.3450108} {Dialect diversity in text summarization on twitter}.
\newblock In \emph{Proceedings of the Web Conference 2021}, WWW '21, page 3802–3814, New York, NY, USA. Association for Computing Machinery.

\bibitem[{Kleiner et~al.(2024)Kleiner, Grieser, Miller, Shepard, Garciv-Perez, Deas, Patton, Turcan, and McKeown}]{kleiner-camouflage}
Shana Kleiner, Jessica~A. Grieser, Shug Miller, James Shepard, Javier Garciv-Perez, Nick Deas, Desmond~U. Patton, Elsbeth Turcan, and Kathleen McKeown. 2024.
\newblock \href {https://doi.org/10.1007/s43681-024-00623-2} {Unmasking camouflage: exploring the challenges of large language models in deciphering african american language and online performativity}.
\newblock \emph{AI and Ethics}.

\bibitem[{Kortmann et~al.(2020)Kortmann, Lunkenheimer, and Ehret}]{kortmann-ewave}
Bernd Kortmann, Kerstin Lunkenheimer, and Katharina Ehret, editors. 2020.
\newblock \href {https://ewave-atlas.org/} {\emph{eWAVE}}.

\bibitem[{Labov et~al.(2011)Labov, Ash, Ravindranath, Weldon, Baranowski, and Nagy}]{labov-monitor}
William Labov, Sharon Ash, Maya Ravindranath, Tracey Weldon, Maciej Baranowski, and Naomi Nagy. 2011.
\newblock \href {https://doi.org/10.1111/j.1467-9841.2011.00504.x} {Properties of the sociolinguistic monitor}.
\newblock \emph{Journal of Sociolinguistics}, 15(4):431--463.

\bibitem[{Ladhak et~al.(2023)Ladhak, Durmus, Suzgun, Zhang, Jurafsky, McKeown, and Hashimoto}]{ladhak-namenat}
Faisal Ladhak, Esin Durmus, Mirac Suzgun, Tianyi Zhang, Dan Jurafsky, Kathleen McKeown, and Tatsunori Hashimoto. 2023.
\newblock \href {https://doi.org/10.18653/v1/2023.eacl-main.234} {When do pre-training biases propagate to downstream tasks? a case study in text summarization}.
\newblock In \emph{Proceedings of the 17th Conference of the European Chapter of the Association for Computational Linguistics}, pages 3206--3219, Dubrovnik, Croatia. Association for Computational Linguistics.

\bibitem[{Lex et~al.(2014)Lex, Gehlenborg, Strobelt, Vuillemot, and Pfister}]{lex-upset}
Alexander Lex, Nils Gehlenborg, Hendrik Strobelt, Romain Vuillemot, and Hanspeter Pfister. 2014.
\newblock \href {https://doi.org/10.1109/tvcg.2014.2346248} {Upset: Visualization of intersecting sets}.
\newblock \emph{IEEE Transactions on Visualization and Computer Graphics}, 20(12):1983–1992.

\bibitem[{Li et~al.(2024)Li, Fang, Ansari, Faghri, Ali, Toshev, Shankar, Smyrnis, Jordan, Igvi, Dimakis, Zhang, Bansal, Vasiljevic, Mercat, Jitsev, Arora, Chen, Muenninghoff, Soldaini, Koh, Heckel, Xin, Gadre, Shao, Pratt, Garg, Keh, Gururangan, Sanyal, Bitton, Kollar, Wortsman, Guha, Abbas, Hsieh, Ghosh, Ilharco, Daras, Marathe, Gardner, Nezhurina, Dave, Carmon, and Schmidt}]{datacomp-lm}
Jeffrey Li, Alex Fang, Hadi~Pour Ansari, Fartash Faghri, Alaaeldin Mohamed~Elnouby Ali, Alexander Toshev, Vaishaal Shankar, Georgios Smyrnis, Matt Jordan, Maor Igvi, Alex Dimakis, Hanlin Zhang, Hritik Bansal, Igor Vasiljevic, Jean Mercat, Jenia Jitsev, Kushal Arora, Mayee Chen, Niklas Muenninghoff, Luca Soldaini, Pang~Wei Koh, Reinhard Heckel, Rui Xin, Samir Gadre, Rulin Shao, Sarah Pratt, Saurabh Garg, Sedrick Keh, Suchin Gururangan, Sunny Sanyal, Yonatan Bitton, Thomas Kollar, Mitchell Wortsman, Etash Guha, Amro Abbas, Cheng-Yu Hsieh, Dhruba Ghosh, Gabriel Ilharco, Giannis Daras, Kalyani Marathe, Joshua Gardner, Marianna Nezhurina, Achal Dave, Yair Carmon, and Ludwig Schmidt. 2024.
\newblock \href {https://arxiv.org/abs/2406.11794} {Datacomp-lm: In search of the next generation of training sets for language models}.

\bibitem[{Liu et~al.(2024)Liu, Min, Zettlemoyer, Choi, and Hajishirzi}]{liu2024infini}
Jiacheng Liu, Sewon Min, Luke Zettlemoyer, Yejin Choi, and Hannaneh Hajishirzi. 2024.
\newblock Infini-gram: Scaling unbounded n-gram language models to a trillion tokens.
\newblock \emph{arXiv preprint arXiv:2401.17377}.

\bibitem[{Lui and Baldwin(2012)}]{lui-langid}
Marco Lui and Timothy Baldwin. 2012.
\newblock \href {https://aclanthology.org/P12-3005} {langid.py: An off-the-shelf language identification tool}.
\newblock In \emph{Proceedings of the {ACL} 2012 System Demonstrations}, pages 25--30, Jeju Island, Korea. Association for Computational Linguistics.

\bibitem[{Manasse et~al.(2010)Manasse, McSherry, and Talwar}]{manasse2010consistent}
Mark Manasse, Frank McSherry, and Kunal Talwar. 2010.
\newblock Consistent weighted sampling.
\newblock \emph{Unpublished technical report) http://research. microsoft. com/en-us/people/manasse}, 2.

\bibitem[{Masis et~al.(2022)Masis, Neal, Green, and O'Connor}]{massis-cgedit}
Tessa Masis, Anissa Neal, Lisa Green, and Brendan O'Connor. 2022.
\newblock \href {https://arxiv.org/abs/2209.07611} {Corpus-guided contrast sets for morphosyntactic feature detection in low-resource english varieties}.
\newblock \emph{Preprint}, arXiv:2209.07611.

\bibitem[{Mire et~al.(2025)Mire, Aysola, Chechelnitsky, Deas, Zerva, and Sap}]{mire-rejected}
Joel Mire, Zubin~Trivadi Aysola, Daniel Chechelnitsky, Nicholas Deas, Chrysoula Zerva, and Maarten Sap. 2025.
\newblock \href {https://aclanthology.org/2025.findings-naacl.417/} {Rejected dialects: Biases against {A}frican {A}merican language in reward models}.
\newblock In \emph{Findings of the Association for Computational Linguistics: NAACL 2025}, pages 7468--7487, Albuquerque, New Mexico. Association for Computational Linguistics.

\bibitem[{Mitchell-Kernan and Thomas(1972)}]{claudia-signifyin}
Claudia Mitchell-Kernan and Kochman Thomas. 1972.
\newblock Signifying, loud-talking and marking.
\newblock \emph{Rappin’and Stylin’Out. Communication in Urban Black America}, pages 315--335.

\bibitem[{Oetting(2015)}]{oetting-swe}
Janna~B. Oetting. 2015.
\newblock \href {https://doi.org/10.1093/oxfordhb/9780199795390.013.23} {{512Some Similarities and Differences Between African American English and Southern White English in Children}}.
\newblock In \emph{{The Oxford Handbook of African American Language}}. Oxford University Press.

\bibitem[{Olabisi et~al.(2022)Olabisi, Hudson, Jetter, and Agrawal}]{olabisi-diversity}
Olubusayo Olabisi, Aaron Hudson, Antonie Jetter, and Ameeta Agrawal. 2022.
\newblock \href {https://aclanthology.org/2022.coling-1.542} {Analyzing the dialect diversity in multi-document summaries}.
\newblock In \emph{Proceedings of the 29th International Conference on Computational Linguistics}, pages 6208--6221, Gyeongju, Republic of Korea. International Committee on Computational Linguistics.

\bibitem[{OLMo et~al.(2024)OLMo, Walsh, Soldaini, Groeneveld, Lo, Arora, Bhagia, Gu, Huang, Jordan, Lambert, Schwenk, Tafjord, Anderson, Atkinson, Brahman, Clark, Dasigi, Dziri, Guerquin, Ivison, Koh, Liu, Malik, Merrill, Miranda, Morrison, Murray, Nam, Pyatkin, Rangapur, Schmitz, Skjonsberg, Wadden, Wilhelm, Wilson, Zettlemoyer, Farhadi, Smith, and Hajishirzi}]{olmo-2}
Team OLMo, Pete Walsh, Luca Soldaini, Dirk Groeneveld, Kyle Lo, Shane Arora, Akshita Bhagia, Yuling Gu, Shengyi Huang, Matt Jordan, Nathan Lambert, Dustin Schwenk, Oyvind Tafjord, Taira Anderson, David Atkinson, Faeze Brahman, Christopher Clark, Pradeep Dasigi, Nouha Dziri, Michal Guerquin, Hamish Ivison, Pang~Wei Koh, Jiacheng Liu, Saumya Malik, William Merrill, Lester James~V. Miranda, Jacob Morrison, Tyler Murray, Crystal Nam, Valentina Pyatkin, Aman Rangapur, Michael Schmitz, Sam Skjonsberg, David Wadden, Christopher Wilhelm, Michael Wilson, Luke Zettlemoyer, Ali Farhadi, Noah~A. Smith, and Hannaneh Hajishirzi. 2024.
\newblock \href {https://arxiv.org/abs/2501.00656} {2 olmo 2 furious}.
\newblock \emph{Preprint}, arXiv:2501.00656.

\bibitem[{OpenAI et~al.(2024)OpenAI, Achiam, Adler, Agarwal, Ahmad, Akkaya, Aleman, Almeida, Altenschmidt, Altman, Anadkat, Avila, Babuschkin, Balaji, Balcom, Baltescu, Bao, Bavarian, Belgum, Bello, Berdine, Bernadett-Shapiro, Berner, Bogdonoff, Boiko, Boyd, Brakman, Brockman, Brooks, Brundage, Button, Cai, Campbell, Cann, Carey, Carlson, Carmichael, Chan, Chang, Chantzis, Chen, Chen, Chen, Chen, Chen, Chess, Cho, Chu, Chung, Cummings, Currier, Dai, Decareaux, Degry, Deutsch, Deville, Dhar, Dohan, Dowling, Dunning, Ecoffet, Eleti, Eloundou, Farhi, Fedus, Felix, Fishman, Forte, Fulford, Gao, Georges, Gibson, Goel, Gogineni, Goh, Gontijo-Lopes, Gordon, Grafstein, Gray, Greene, Gross, Gu, Guo, Hallacy, Han, Harris, He, Heaton, Heidecke, Hesse, Hickey, Hickey, Hoeschele, Houghton, Hsu, Hu, Hu, Huizinga, Jain, Jain, Jang, Jiang, Jiang, Jin, Jin, Jomoto, Jonn, Jun, Kaftan, Łukasz Kaiser, Kamali, Kanitscheider, Keskar, Khan, Kilpatrick, Kim, Kim, Kim, Kirchner, Kiros, Knight, Kokotajlo, Łukasz Kondraciuk,
  Kondrich, Konstantinidis, Kosic, Krueger, Kuo, Lampe, Lan, Lee, Leike, Leung, Levy, Li, Lim, Lin, Lin, Litwin, Lopez, Lowe, Lue, Makanju, Malfacini, Manning, Markov, Markovski, Martin, Mayer, Mayne, McGrew, McKinney, McLeavey, McMillan, McNeil, Medina, Mehta, Menick, Metz, Mishchenko, Mishkin, Monaco, Morikawa, Mossing, Mu, Murati, Murk, Mély, Nair, Nakano, Nayak, Neelakantan, Ngo, Noh, Ouyang, O'Keefe, Pachocki, Paino, Palermo, Pantuliano, Parascandolo, Parish, Parparita, Passos, Pavlov, Peng, Perelman, de~Avila Belbute~Peres, Petrov, de~Oliveira~Pinto, Michael, Pokorny, Pokrass, Pong, Powell, Power, Power, Proehl, Puri, Radford, Rae, Ramesh, Raymond, Real, Rimbach, Ross, Rotsted, Roussez, Ryder, Saltarelli, Sanders, Santurkar, Sastry, Schmidt, Schnurr, Schulman, Selsam, Sheppard, Sherbakov, Shieh, Shoker, Shyam, Sidor, Sigler, Simens, Sitkin, Slama, Sohl, Sokolowsky, Song, Staudacher, Such, Summers, Sutskever, Tang, Tezak, Thompson, Tillet, Tootoonchian, Tseng, Tuggle, Turley, Tworek, Uribe, Vallone,
  Vijayvergiya, Voss, Wainwright, Wang, Wang, Wang, Ward, Wei, Weinmann, Welihinda, Welinder, Weng, Weng, Wiethoff, Willner, Winter, Wolrich, Wong, Workman, Wu, Wu, Wu, Xiao, Xu, Yoo, Yu, Yuan, Zaremba, Zellers, Zhang, Zhang, Zhao, Zheng, Zhuang, Zhuk, and Zoph}]{gpt4}
OpenAI, Josh Achiam, Steven Adler, Sandhini Agarwal, Lama Ahmad, Ilge Akkaya, Florencia~Leoni Aleman, Diogo Almeida, Janko Altenschmidt, Sam Altman, Shyamal Anadkat, Red Avila, Igor Babuschkin, Suchir Balaji, Valerie Balcom, Paul Baltescu, Haiming Bao, Mohammad Bavarian, Jeff Belgum, Irwan Bello, Jake Berdine, Gabriel Bernadett-Shapiro, Christopher Berner, Lenny Bogdonoff, Oleg Boiko, Madelaine Boyd, Anna-Luisa Brakman, Greg Brockman, Tim Brooks, Miles Brundage, Kevin Button, Trevor Cai, Rosie Campbell, Andrew Cann, Brittany Carey, Chelsea Carlson, Rory Carmichael, Brooke Chan, Che Chang, Fotis Chantzis, Derek Chen, Sully Chen, Ruby Chen, Jason Chen, Mark Chen, Ben Chess, Chester Cho, Casey Chu, Hyung~Won Chung, Dave Cummings, Jeremiah Currier, Yunxing Dai, Cory Decareaux, Thomas Degry, Noah Deutsch, Damien Deville, Arka Dhar, David Dohan, Steve Dowling, Sheila Dunning, Adrien Ecoffet, Atty Eleti, Tyna Eloundou, David Farhi, Liam Fedus, Niko Felix, Simón~Posada Fishman, Juston Forte, Isabella Fulford, Leo
  Gao, Elie Georges, Christian Gibson, Vik Goel, Tarun Gogineni, Gabriel Goh, Rapha Gontijo-Lopes, Jonathan Gordon, Morgan Grafstein, Scott Gray, Ryan Greene, Joshua Gross, Shixiang~Shane Gu, Yufei Guo, Chris Hallacy, Jesse Han, Jeff Harris, Yuchen He, Mike Heaton, Johannes Heidecke, Chris Hesse, Alan Hickey, Wade Hickey, Peter Hoeschele, Brandon Houghton, Kenny Hsu, Shengli Hu, Xin Hu, Joost Huizinga, Shantanu Jain, Shawn Jain, Joanne Jang, Angela Jiang, Roger Jiang, Haozhun Jin, Denny Jin, Shino Jomoto, Billie Jonn, Heewoo Jun, Tomer Kaftan, Łukasz Kaiser, Ali Kamali, Ingmar Kanitscheider, Nitish~Shirish Keskar, Tabarak Khan, Logan Kilpatrick, Jong~Wook Kim, Christina Kim, Yongjik Kim, Jan~Hendrik Kirchner, Jamie Kiros, Matt Knight, Daniel Kokotajlo, Łukasz Kondraciuk, Andrew Kondrich, Aris Konstantinidis, Kyle Kosic, Gretchen Krueger, Vishal Kuo, Michael Lampe, Ikai Lan, Teddy Lee, Jan Leike, Jade Leung, Daniel Levy, Chak~Ming Li, Rachel Lim, Molly Lin, Stephanie Lin, Mateusz Litwin, Theresa Lopez, Ryan
  Lowe, Patricia Lue, Anna Makanju, Kim Malfacini, Sam Manning, Todor Markov, Yaniv Markovski, Bianca Martin, Katie Mayer, Andrew Mayne, Bob McGrew, Scott~Mayer McKinney, Christine McLeavey, Paul McMillan, Jake McNeil, David Medina, Aalok Mehta, Jacob Menick, Luke Metz, Andrey Mishchenko, Pamela Mishkin, Vinnie Monaco, Evan Morikawa, Daniel Mossing, Tong Mu, Mira Murati, Oleg Murk, David Mély, Ashvin Nair, Reiichiro Nakano, Rajeev Nayak, Arvind Neelakantan, Richard Ngo, Hyeonwoo Noh, Long Ouyang, Cullen O'Keefe, Jakub Pachocki, Alex Paino, Joe Palermo, Ashley Pantuliano, Giambattista Parascandolo, Joel Parish, Emy Parparita, Alex Passos, Mikhail Pavlov, Andrew Peng, Adam Perelman, Filipe de~Avila Belbute~Peres, Michael Petrov, Henrique~Ponde de~Oliveira~Pinto, Michael, Pokorny, Michelle Pokrass, Vitchyr~H. Pong, Tolly Powell, Alethea Power, Boris Power, Elizabeth Proehl, Raul Puri, Alec Radford, Jack Rae, Aditya Ramesh, Cameron Raymond, Francis Real, Kendra Rimbach, Carl Ross, Bob Rotsted, Henri Roussez,
  Nick Ryder, Mario Saltarelli, Ted Sanders, Shibani Santurkar, Girish Sastry, Heather Schmidt, David Schnurr, John Schulman, Daniel Selsam, Kyla Sheppard, Toki Sherbakov, Jessica Shieh, Sarah Shoker, Pranav Shyam, Szymon Sidor, Eric Sigler, Maddie Simens, Jordan Sitkin, Katarina Slama, Ian Sohl, Benjamin Sokolowsky, Yang Song, Natalie Staudacher, Felipe~Petroski Such, Natalie Summers, Ilya Sutskever, Jie Tang, Nikolas Tezak, Madeleine~B. Thompson, Phil Tillet, Amin Tootoonchian, Elizabeth Tseng, Preston Tuggle, Nick Turley, Jerry Tworek, Juan Felipe~Cerón Uribe, Andrea Vallone, Arun Vijayvergiya, Chelsea Voss, Carroll Wainwright, Justin~Jay Wang, Alvin Wang, Ben Wang, Jonathan Ward, Jason Wei, CJ~Weinmann, Akila Welihinda, Peter Welinder, Jiayi Weng, Lilian Weng, Matt Wiethoff, Dave Willner, Clemens Winter, Samuel Wolrich, Hannah Wong, Lauren Workman, Sherwin Wu, Jeff Wu, Michael Wu, Kai Xiao, Tao Xu, Sarah Yoo, Kevin Yu, Qiming Yuan, Wojciech Zaremba, Rowan Zellers, Chong Zhang, Marvin Zhang, Shengjia
  Zhao, Tianhao Zheng, Juntang Zhuang, William Zhuk, and Barret Zoph. 2024.
\newblock \href {https://arxiv.org/abs/2303.08774} {Gpt-4 technical report}.
\newblock \emph{Preprint}, arXiv:2303.08774.

\bibitem[{pandas~development team(2020)}]{reback2020pandas}
The pandas~development team. 2020.
\newblock \href {https://doi.org/10.5281/zenodo.3509134} {pandas-dev/pandas: Pandas}.

\bibitem[{Patton et~al.(2020)Patton, Frey, McGregor, Lee, McKeown, and Moss}]{patton-casm}
Desmond~U. Patton, William~R. Frey, Kyle~A. McGregor, Fei-Tzin Lee, Kathleen McKeown, and Emanuel Moss. 2020.
\newblock \href {https://doi.org/10.1145/3375627.3375841} {Contextual analysis of social media: The promise and challenge of eliciting context in social media posts with natural language processing}.
\newblock In \emph{Proceedings of the AAAI/ACM Conference on AI, Ethics, and Society}, AIES '20, page 337–342, New York, NY, USA. Association for Computing Machinery.

\bibitem[{Penedo et~al.(2024)Penedo, Kydl{\'\i}{\v{c}}ek, allal, Lozhkov, Mitchell, Raffel, Werra, and Wolf}]{fineweb}
Guilherme Penedo, Hynek Kydl{\'\i}{\v{c}}ek, Loubna~Ben allal, Anton Lozhkov, Margaret Mitchell, Colin Raffel, Leandro~Von Werra, and Thomas Wolf. 2024.
\newblock \href {https://openreview.net/forum?id=n6SCkn2QaG} {The fineweb datasets: Decanting the web for the finest text data at scale}.
\newblock In \emph{The Thirty-eight Conference on Neural Information Processing Systems Datasets and Benchmarks Track}.

\bibitem[{Penedo et~al.(2023)Penedo, Malartic, Hesslow, Cojocaru, Alobeidli, Cappelli, Pannier, Almazrouei, and Launay}]{refinedweb}
Guilherme Penedo, Quentin Malartic, Daniel Hesslow, Ruxandra Cojocaru, Hamza Alobeidli, Alessandro Cappelli, Baptiste Pannier, Ebtesam Almazrouei, and Julien Launay. 2023.
\newblock \href {https://proceedings.neurips.cc/paper_files/paper/2023/file/fa3ed726cc5073b9c31e3e49a807789c-Paper-Datasets_and_Benchmarks.pdf} {The refinedweb dataset for falcon llm: Outperforming curated corpora with web data only}.
\newblock In \emph{Advances in Neural Information Processing Systems}, volume~36, pages 79155--79172. Curran Associates, Inc.

\bibitem[{Raffel et~al.(2020)Raffel, Shazeer, Roberts, Lee, Narang, Matena, Zhou, Li, and Liu}]{raffel-t5}
Colin Raffel, Noam Shazeer, Adam Roberts, Katherine Lee, Sharan Narang, Michael Matena, Yanqi Zhou, Wei Li, and Peter~J Liu. 2020.
\newblock Exploring the limits of transfer learning with a unified text-to-text transformer.
\newblock \emph{The Journal of Machine Learning Research}, 21(1):5485--5551.

\bibitem[{Ronkin and Karn(1999)}]{ronkin-mock}
Maggie Ronkin and Helen~E Karn. 1999.
\newblock Mock ebonics: Linguistic racism in parodies of ebonics on the internet.
\newblock \emph{Journal of sociolinguistics}, 3(3):360--380.

\bibitem[{Roth-Gordon et~al.(2020)Roth-Gordon, Harris, and Zamora}]{gordon-comfort}
Jennifer Roth-Gordon, Jessica Harris, and Stephanie Zamora. 2020.
\newblock \href {https://doi.org/10.1515/ijsl-2020-2105} {Producing white comfort through “corporate cool”: Linguistic appropriation, social media, and @brandssayingbae}.
\newblock \emph{International Journal of the Sociology of Language}, 2020(265):107–128.

\bibitem[{Sachdeva et~al.(2024)Sachdeva, Coleman, Kang, Ni, Hong, Chi, Caverlee, McAuley, and Cheng}]{sachdeva-askllm}
Noveen Sachdeva, Benjamin Coleman, Wang-Cheng Kang, Jianmo Ni, Lichan Hong, Ed~H. Chi, James Caverlee, Julian McAuley, and Derek~Zhiyuan Cheng. 2024.
\newblock \href {https://arxiv.org/abs/2402.09668} {How to train data-efficient llms}.
\newblock \emph{Preprint}, arXiv:2402.09668.

\bibitem[{Sap et~al.(2019{\natexlab{a}})Sap, Card, Gabriel, Choi, and Smith}]{sap-bias}
Maarten Sap, Dallas Card, Saadia Gabriel, Yejin Choi, and Noah~A. Smith. 2019{\natexlab{a}}.
\newblock \href {https://doi.org/10.18653/v1/P19-1163} {The risk of racial bias in hate speech detection}.
\newblock In \emph{Proceedings of the 57th Annual Meeting of the Association for Computational Linguistics}, pages 1668--1678, Florence, Italy. Association for Computational Linguistics.

\bibitem[{Sap et~al.(2019{\natexlab{b}})Sap, Card, Gabriel, Choi, and Smith}]{sap-hatespeech}
Maarten Sap, Dallas Card, Saadia Gabriel, Yejin Choi, and Noah~A. Smith. 2019{\natexlab{b}}.
\newblock \href {https://doi.org/10.18653/v1/P19-1163} {The risk of racial bias in hate speech detection}.
\newblock In \emph{Proceedings of the 57th Annual Meeting of the Association for Computational Linguistics}, pages 1668--1678, Florence, Italy. Association for Computational Linguistics.

\bibitem[{Sap et~al.(2022)Sap, Swayamdipta, Vianna, Zhou, Choi, and Smith}]{sap-annot}
Maarten Sap, Swabha Swayamdipta, Laura Vianna, Xuhui Zhou, Yejin Choi, and Noah~A. Smith. 2022.
\newblock \href {https://doi.org/10.18653/v1/2022.naacl-main.431} {Annotators with attitudes: How annotator beliefs and identities bias toxic language detection}.
\newblock In \emph{Proceedings of the 2022 Conference of the North American Chapter of the Association for Computational Linguistics: Human Language Technologies}, pages 5884--5906, Seattle, United States. Association for Computational Linguistics.

\bibitem[{Soldaini et~al.(2024)Soldaini, Kinney, Bhagia, Schwenk, Atkinson, Authur, Bogin, Chandu, Dumas, Elazar, Hofmann, Jha, Kumar, Lucy, Lyu, Lambert, Magnusson, Morrison, Muennighoff, Naik, Nam, Peters, Ravichander, Richardson, Shen, Strubell, Subramani, Tafjord, Walsh, Zettlemoyer, Smith, Hajishirzi, Beltagy, Groeneveld, Dodge, and Lo}]{soldaini-dolma}
Luca Soldaini, Rodney Kinney, Akshita Bhagia, Dustin Schwenk, David Atkinson, Russell Authur, Ben Bogin, Khyathi Chandu, Jennifer Dumas, Yanai Elazar, Valentin Hofmann, Ananya Jha, Sachin Kumar, Li~Lucy, Xinxi Lyu, Nathan Lambert, Ian Magnusson, Jacob Morrison, Niklas Muennighoff, Aakanksha Naik, Crystal Nam, Matthew Peters, Abhilasha Ravichander, Kyle Richardson, Zejiang Shen, Emma Strubell, Nishant Subramani, Oyvind Tafjord, Evan Walsh, Luke Zettlemoyer, Noah Smith, Hannaneh Hajishirzi, Iz~Beltagy, Dirk Groeneveld, Jesse Dodge, and Kyle Lo. 2024.
\newblock \href {https://aclanthology.org/2024.acl-long.840} {Dolma: an open corpus of three trillion tokens for language model pretraining research}.
\newblock In \emph{Proceedings of the 62nd Annual Meeting of the Association for Computational Linguistics (Volume 1: Long Papers)}, pages 15725--15788, Bangkok, Thailand. Association for Computational Linguistics.

\bibitem[{Suresh et~al.(2024)Suresh, Tseng, Young, Gray, Pierson, and Levy}]{suresh-foundation}
Harini Suresh, Emily Tseng, Meg Young, Mary Gray, Emma Pierson, and Karen Levy. 2024.
\newblock \href {https://doi.org/10.1145/3630106.3658992} {Participation in the age of foundation models}.
\newblock In \emph{Proceedings of the 2024 ACM Conference on Fairness, Accountability, and Transparency}, FAccT '24, page 1609–1621, New York, NY, USA. Association for Computing Machinery.

\bibitem[{Vink et~al.(2025)Vink, Welling, Shima, and Bacher}]{polarsguide}
Ritchie Vink, Soren Welling, Tatsuya Shima, and Etienne Bacher. 2025.
\newblock \href {https://github.com/pola-rs/r-polars} {\emph{polars: Lightning-Fast 'DataFrame' Library}}.
\newblock R package version 0.22.0,.

\bibitem[{Weber et~al.(2024)Weber, Fu, Anthony, Oren, Adams, Alexandrov, Lyu, Nguyen, Yao, Adams, Athiwaratkun, Chalamala, Chen, Ryabinin, Dao, Liang, Ré, Rish, and Zhang}]{redpajama-v2}
Maurice Weber, Daniel~Y. Fu, Quentin Anthony, Yonatan Oren, Shane Adams, Anton Alexandrov, Xiaozhong Lyu, Huu Nguyen, Xiaozhe Yao, Virginia Adams, Ben Athiwaratkun, Rahul Chalamala, Kezhen Chen, Max Ryabinin, Tri Dao, Percy Liang, Christopher Ré, Irina Rish, and Ce~Zhang. 2024.
\newblock Redpajama: an open dataset for training large language models.
\newblock \emph{NeurIPS Datasets and Benchmarks Track}.

\bibitem[{Wettig et~al.(2024)Wettig, Gupta, Malik, and Chen}]{wettig-qurating}
Alexander Wettig, Aatmik Gupta, Saumya Malik, and Danqi Chen. 2024.
\newblock \href {https://proceedings.mlr.press/v235/wettig24a.html} {{Q}u{R}ating: Selecting high-quality data for training language models}.
\newblock In \emph{Proceedings of the 41st International Conference on Machine Learning}, volume 235 of \emph{Proceedings of Machine Learning Research}, pages 52915--52971. PMLR.

\bibitem[{Wolfram and Kohn(2015)}]{wolfram2015regionality}
Walt Wolfram and Mary~E Kohn. 2015.
\newblock Regionality in the development of african american english.
\newblock \emph{The Oxford handbook of African American language}, pages 140--160.

\bibitem[{Xia et~al.(2020)Xia, Field, and Tsvetkov}]{xia-demoting}
Mengzhou Xia, Anjalie Field, and Yulia Tsvetkov. 2020.
\newblock \href {https://doi.org/10.18653/v1/2020.socialnlp-1.2} {Demoting racial bias in hate speech detection}.
\newblock In \emph{Proceedings of the Eighth International Workshop on Natural Language Processing for Social Media}, pages 7--14, Online. Association for Computational Linguistics.

\bibitem[{Zellers et~al.(2018)Zellers, Bisk, Schwartz, and Choi}]{zellers-swag}
Rowan Zellers, Yonatan Bisk, Roy Schwartz, and Yejin Choi. 2018.
\newblock \href {https://doi.org/10.18653/v1/D18-1009} {{SWAG}: A large-scale adversarial dataset for grounded commonsense inference}.
\newblock In \emph{Proceedings of the 2018 Conference on Empirical Methods in Natural Language Processing}, pages 93--104, Brussels, Belgium. Association for Computational Linguistics.

\bibitem[{Zellers et~al.(2019)Zellers, Holtzman, Bisk, Farhadi, and Choi}]{zellers-hellaswag}
Rowan Zellers, Ari Holtzman, Yonatan Bisk, Ali Farhadi, and Yejin Choi. 2019.
\newblock \href {https://doi.org/10.18653/v1/P19-1472} {{H}ella{S}wag: Can a machine really finish your sentence?}
\newblock In \emph{Proceedings of the 57th Annual Meeting of the Association for Computational Linguistics}, pages 4791--4800, Florence, Italy. Association for Computational Linguistics.

\bibitem[{Ziems et~al.(2022)Ziems, Chen, Harris, Anderson, and Yang}]{ziems-value}
Caleb Ziems, Jiaao Chen, Camille Harris, Jessica Anderson, and Diyi Yang. 2022.
\newblock \href {https://doi.org/10.18653/v1/2022.acl-long.258} {{VALUE}: {U}nderstanding dialect disparity in {NLU}}.
\newblock In \emph{Proceedings of the 60th Annual Meeting of the Association for Computational Linguistics (Volume 1: Long Papers)}, pages 3701--3720, Dublin, Ireland. Association for Computational Linguistics.

\end{thebibliography}

\appendix

\section{Large-Scale Corpora Analysis}
\label{app:corpus-dets}

We select \numcorpora{} open-source pretraining corpora for our analyses, prioritizing those that have been used to pretrain models often used in academic research. Given our focus on AAL, we also focus analyses  on corpora that are intended for pretraining predominantly English rather than multilingual models. For all corpora, we strictly analyze portions that are publicly available and do not include subsets of RedPajama \cite{redpajama} or The Pile \cite{thepile} that are restricted due to copyright. All corpora are only used for academic research purposes in line with their intent. 

\subsection{Pretraining Corpus Scanning}
\label{app:corpus-full}

    Our data pipeline is written in Python with Polars \cite{polarsguide} for performance-sensitive data processing, Pandas \cite{reback2020pandas}, UpSetPlot \cite{lex-upset}, and MatPlotlib \cite{Hunter:2007} for analyses and visualization. Labeling of independent subsets of the data is a trivially parallelizable task. We partition the corpus into disjoint splits and use 20 independent worker processes, one for each available physical CPU core to store the full text of records matching our selection criterion. In all, we obtain labels for over 16 TB of compressed records, as stored on disk. For capturing approximations of full distributions such as in \ref{fig:prob-dist}, we additionally use $t$-digest \cite{tdigest}.
    
\subsection{Corpus Sample Analyses}
\label{app:corpus-sample}

 Our file sampling policy was motivated by the constraints of the available metadata, the size of the corpora, and the goal of our analysis being reproducible. File sizes vary between corpora and within corpora, so a fixed number of files would not result in balanced representation. Due to the low metadata consistency of the corpora, the only common attribute to all corpora was the URL hosting the data file. To create a uniform random sampling of the files in the corpora, we use a variant on hash-based consistent sampling \cite{manasse2010consistent}. Thus, the selection procedure entailed hashing the full url hosting the data with sha256, then sampling the first $n$ files (sorted by hash) until a threshold of 250 GB was reached for each corpus.

\begin{figure*}
\section{Additional Data Details} %
\label{app:data-det}
    \begin{center}
   \resizebox{0.96\textwidth}{!}{\includegraphics{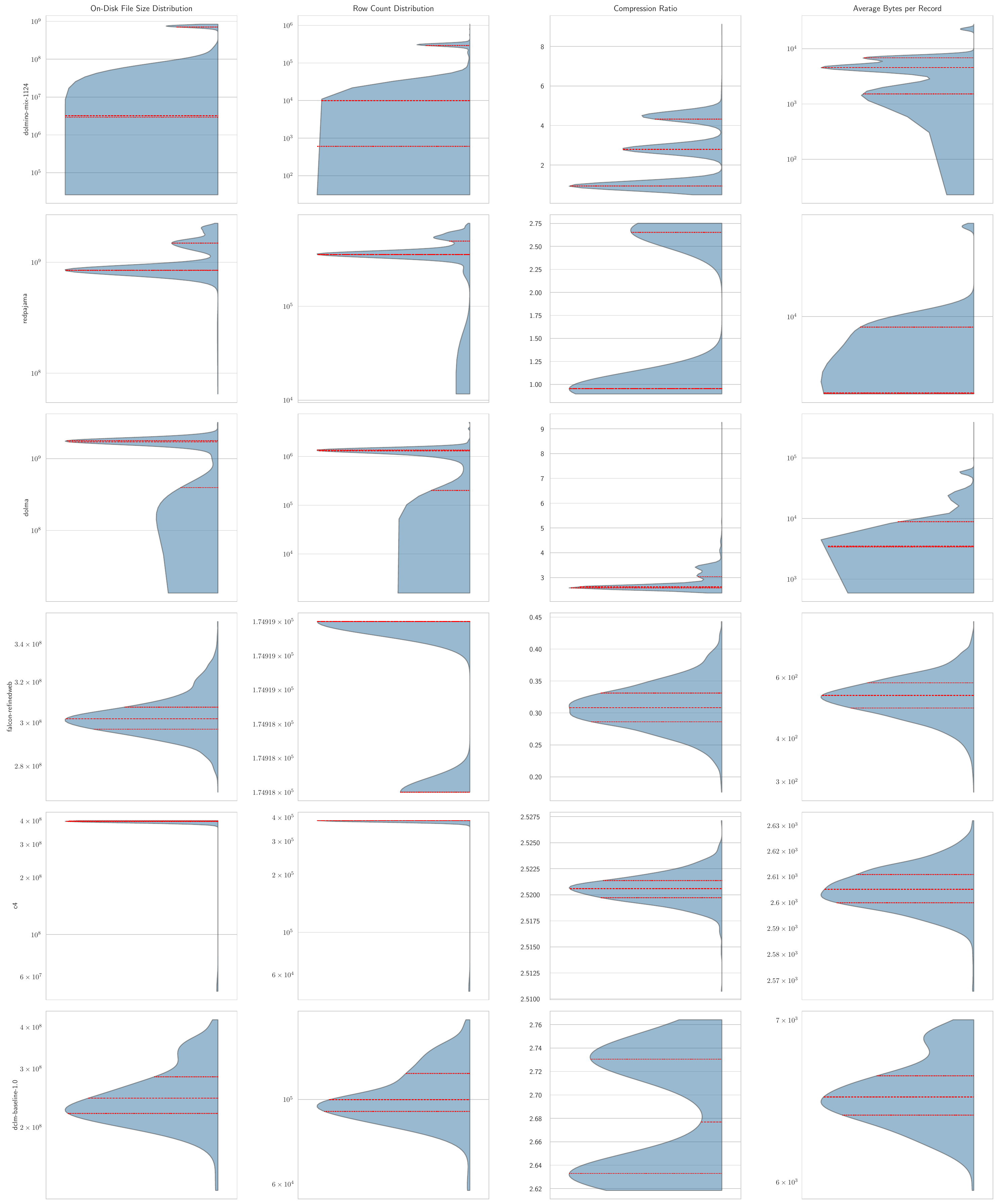}} 
    \end{center}
    \caption{Distribution of row count, compression ratio, file size, and bytes per record from select corpora. This captures the variety of metadata standards between the different corpora. 
    }
\end{figure*}

\section{AAL Probability Threshold}
\label{app:threshold}

\begin{figure}
    \begin{center}
    \resizebox{0.4\textwidth}{!}{\includegraphics{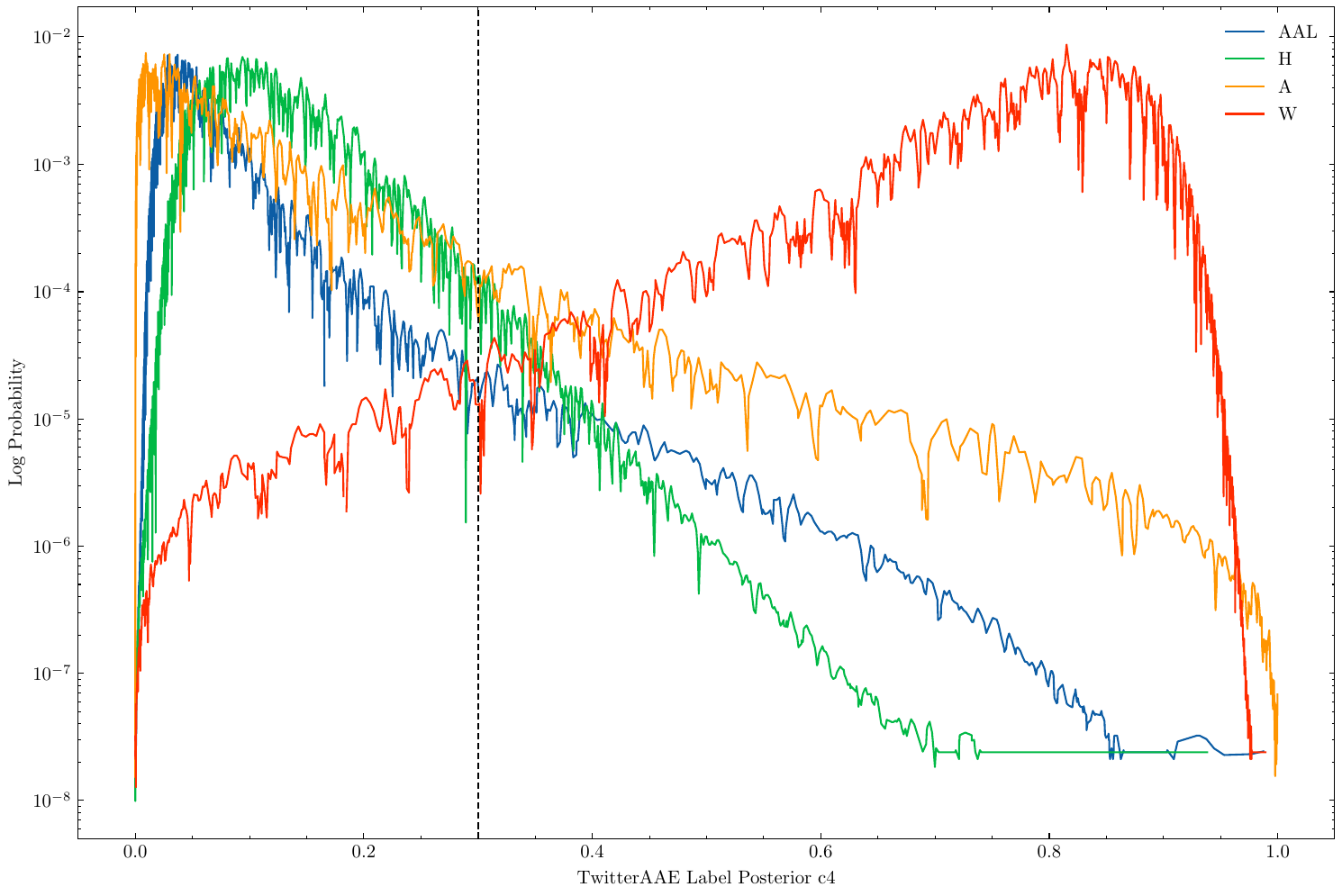}}
    \resizebox{0.4\textwidth}{!}{\includegraphics{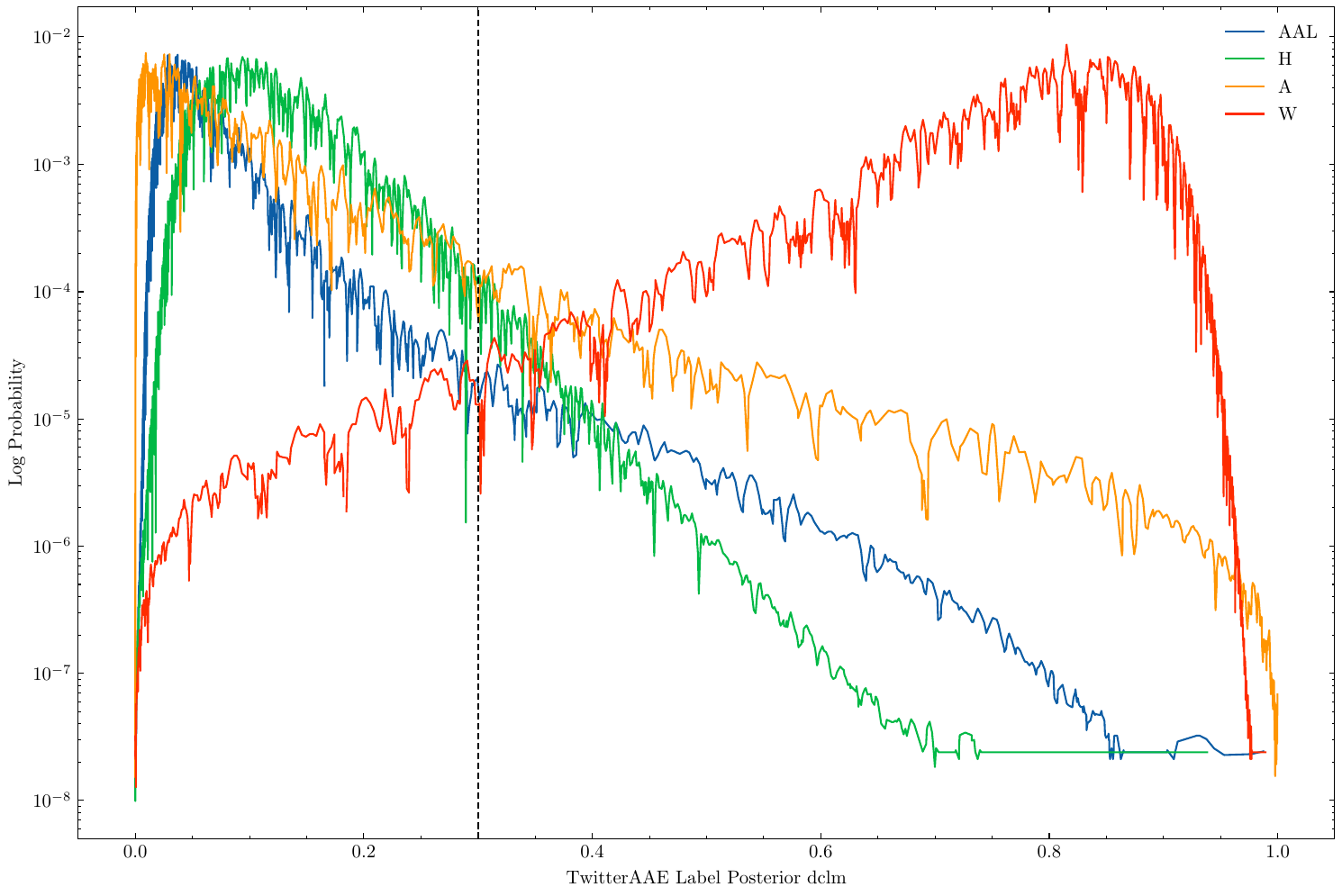}} 
    \resizebox{0.4\textwidth}{!}{\includegraphics{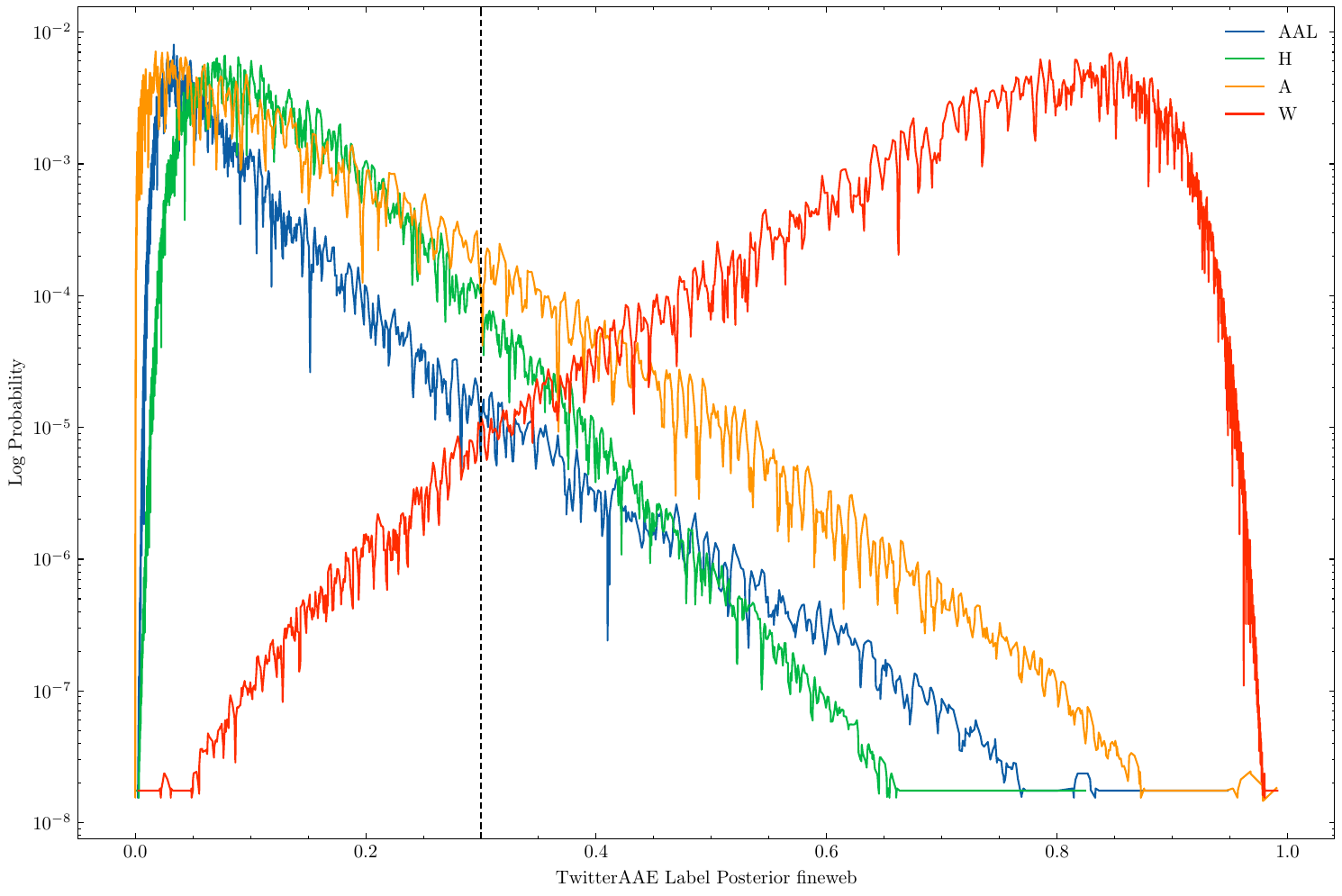}} 
    \resizebox{0.4\textwidth}{!}{\includegraphics{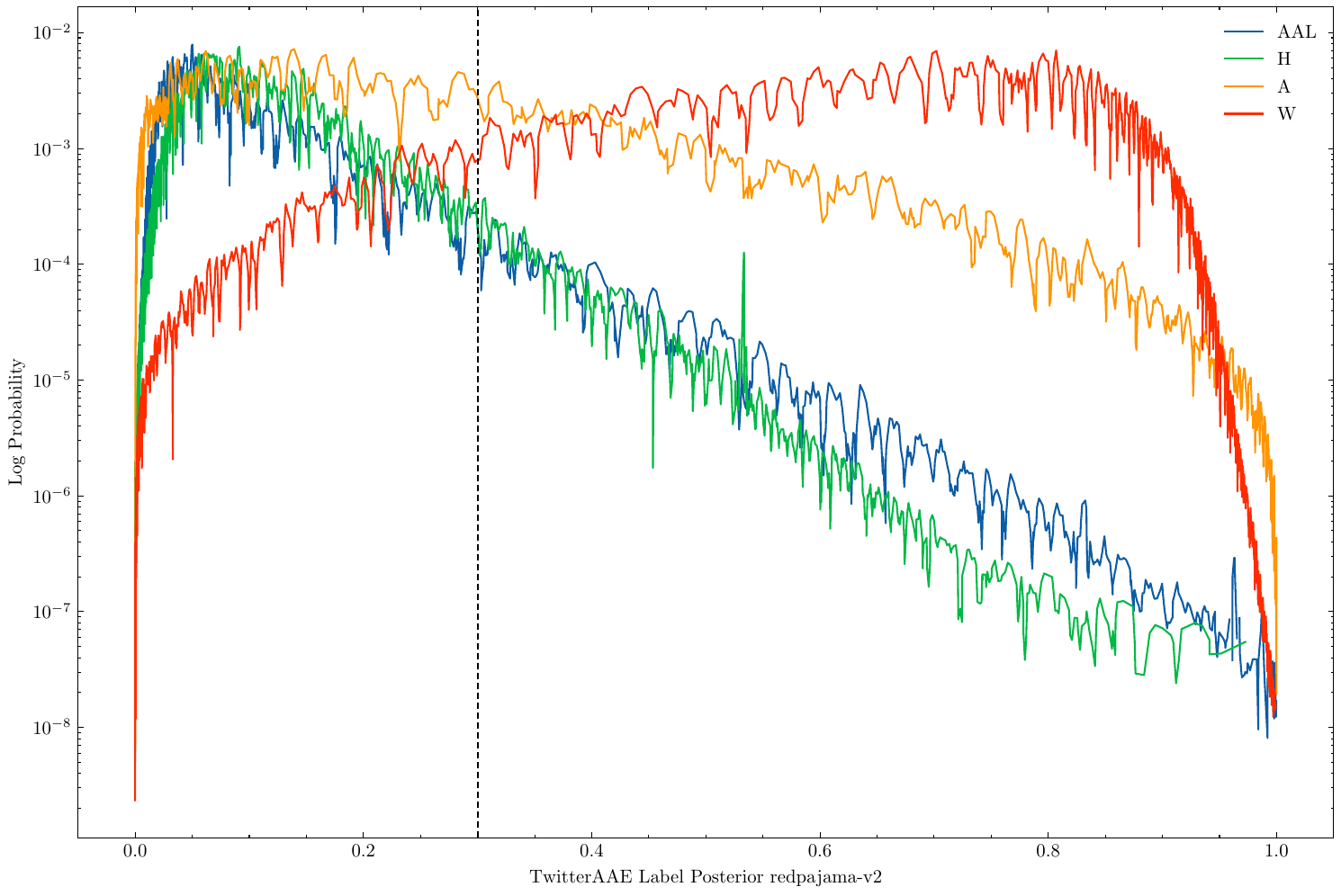}}  
    \end{center}
    \caption{Full distribution of demographic alignment classifier posterior probabilities across all TwitterAAE Model Labels for C4 and all sampled corpora DCLM-baseline, RedPajama-Data-V2, and FineWeb, as captured by t-digest \protect\cite{tdigest}.
    }
    \label{fig:prob-dist}
\end{figure}

Any record where AAL has the greatest probability of all labels or where AAL exceeds a posterior probability of 0.3 is considered for analysis. This threshold is chosen for the individual feature analyses to capture documents that may contain features of AAL although largely be composed of WME. At the same time, this threshold ensures that the  
extracted dataset is feasible to analyze with the available computational resources. \autoref{fig:prob-dist} shows the distribution of document AAL probabilities as well as the probabilities of the other language varieties considered by the demographic alignment classifier \cite{blodgett-variation}. For C4 (top), considering lower thresholds sharply increases the number of documents considered.

\section{Human Judgments}
\label{app:judgments}

\subsection{Annotation Details}

We recruit 3 annotators that self-identify as AAL speakers and have previous or current experience in natural language processing or linguistics. Each annotator is initially provided 420 texts total, with 50 texts shared among all annotators. Two annotators completed judgments for all 420 texts, and the remaining annotator completed 299 texts. Annotators were compensated at a rate of \$22.50 per hour.

\subsection{Human Judgments Sample}

Because much of recent work relies on a threshold of 0.8 to identify AAL documents (e.g., \citet{davidson-datasets}) using the demographic alignment classifier \cite{blodgett-variation},
we prioritize texts with a higher probability of including AAL in human judgments. Additionally, because \unfiltered{} is not used for pretraining, we prioritize texts from the filtered subset of C4 (\filtered{}). The composition enforced on the sampled human judgment texts is shown in \autoref{tab:annot-sample}. Notably, only 10\% of the texts in the sample have $\geq90\%$ probability because there are few texts in the full extracted dataset that meet this requirement.
Considering this sampling, all reported estimates based on human judgments (e.g., overall \textit{Stereotype} estimates) are calculated through a weighted average across probability ranges, weighted by the size of each probability range in the full corpus. 

\begin{table}[!htbp]
            \centering
            \small
            \begin{tabular}{| c | c |}
                \hline
                \textbf{AA Prob} & \% \textbf{Sample} \\
                \hline
                $.9 \leq P_{AA}$ & 0.1  \\
                $.8 \geq P_{AA} \geq .9$ & 0.35 \\
                $.7 \geq P_{AA} \geq .8$ & 0.3  \\
                $.6 \geq P_{AA} \geq .7$ & 0.15 \\
                $.5 \geq P_{AA} \geq .6$ & 0.1 \\
                \hline
                \hline
                \textbf{Filtering} & \% \textbf{Sample} \\
                \hline
                Filtered (\filtered{}) & 0.75 \\
                Unfiltered (\unfiltered{}) & 0.25 \\
                \hline
            \end{tabular}
            \caption{Percent sampled from each subset of the extracted AAL C4 data for human judgments. }
            \label{tab:annot-sample}
        \end{table}

\subsection{Annotator Agreement}

\begin{table}[!htbp]
    \centering
    \small
    \begin{tabular}{| c | c c c | c |}
        \hline
        Dimension           &  $\kappa$         &  $\kappa_{bin}$           &  $r_s$        & Support \\
        \hline
        Human               &  0.226            &  0.581                    & 0.588         & \multirow{3}{*}{121} \\
        Dialect             &  0.357            &  0.747                    & 0.699         & \\
        Native              &  0.095            &  0.619                    & 0.572         & \\
        \hline
        Approp              &  0.101            &  0.188                    & 0.336         & \multirow{2}{*}{41} \\
        Stereo              &  0.009            & -0.021                    & 0.083         & \\
        \hline
        \end{tabular}
    \caption{Average pairwise interannotator agreement results for each dimension using Cohen's $\kappa$ and Spearman $r_s$. Cohen's $\kappa_{bin}$ represents agreement with binarized judgments ($\{0,1\}\rightarrow0$, $\{2,3\}\rightarrow1$).}
    \label{tab:annot-agree}
\end{table}

\autoref{tab:annot-agree} presents the inter-annotator agreement scores for each judgment dimension. Annotators show moderate to substantial agreement for binarized \textit{Humanness} ($\kappa_{bin} = 0.581$), \textit{Dialect} ($\kappa_{bin} = 0.747$), and \textit{Native Speaker} ($\kappa_{bin} = 0.619$), as well as strong Spearman correlations.

Agreements on the two remaining perceptual dimensions, \textit{Appropriateness} and \textit{Stereotype}, are expectedly low considering that annotators personal linguistic attitudes can lead to varying interpretations of what constitutes "appropriate" or appears to reinforce stereotypes. However, annotators exhibit slight agreement on \textit{Appropriateness}.

\subsection{Annotation Interface}
\label{app:annot-inter}

\begin{figure*}[!htbp]
    \centering
    \includegraphics[width=0.9\linewidth]{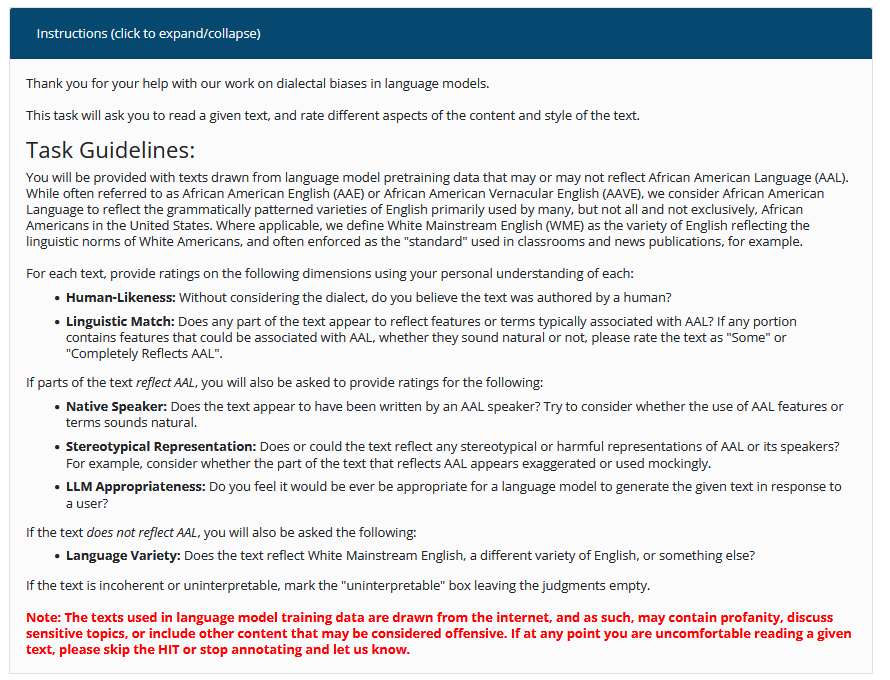}
    \caption{Screenshot of annotation interface instructions panel.}
    \label{fig:interface-1}
\end{figure*}

\begin{figure*}[!htbp]
    \centering
    \includegraphics[width=0.9\linewidth]{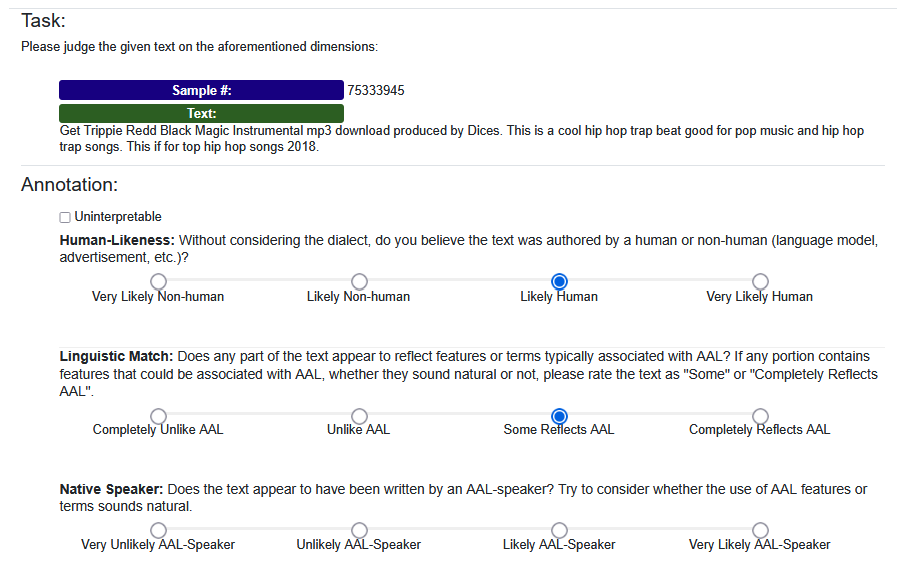}
    \caption{Screenshot of \textit{Human-Likeness}, \textit{Linguistic Match}, and \textit{Native Speaker} questions in annotation interface.}
    \label{fig:interface-2}
\end{figure*}

\begin{figure*}[!htbp]
    \centering
    \includegraphics[width=0.9\linewidth]{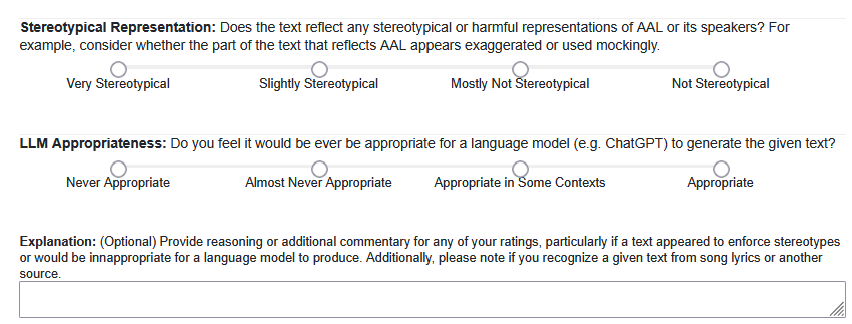}
    \caption{Screenshot of \textit{Stereotype} and \textit{Appropriateness} questions in annotation interface.}
    \label{fig:interface-3}
\end{figure*}

We also include an additional space for comments, including reasoning for selected judgments as well as if the annotator recognizes a given text from a different source (e.g., song lyrics).
Screenshots of the interface are provided in \autoref{fig:interface-1}, \autoref{fig:interface-2}, and \autoref{fig:interface-3}.

\section{Feature Abbreviations}
\label{app:feat-explain}

In analyses of morphosyntactic features, we consider the features listed in \citet{massis-cgedit}. Abbreviations used for each feature are included in \autoref{tab:feat-map}. 

\begin{table}[!htbp]
    \centering
    \small
    \begin{tabular}{| c | c |}
        \hline
        Abbrev. & Feature                               \\
        \hline
        ZP      & Zero Possessive                                 \\
        ZC      & Zero Copula                                     \\
        DT      & Double Tense                                    \\
        HB      & Habitual Be                                     \\
        RD      & Resultant/Completive \textit{Done}              \\
        FINNA   & \textit{finna}                                  \\
        COME    & \textit{come}                                   \\
        DM      & Double Modal                                    \\
        MN      & Multiple Negation                               \\
        NAI     & Negative Auxiliary Inversion                    \\
        NINC    & Non-Inverted Negative Concord                   \\
        AI      & \textit{ain't}                                  \\
        3S      & Zero 3rd-Person Singular Present \textit{-s}    \\
        IW      & \textit{is/was} Generalization                  \\
        ZPL     & Zero Plural \textit{-s}                         \\
        DO      & Double Object                                   \\
        WH      & \textit{Wh-} Question                           \\
        \hline
    \end{tabular}
    \caption{Feature abbreviations and names covered by the CGEdit model  \protect\cite{massis-cgedit}. Examples are identified from pretraining corpora using model predictions.}
    \label{tab:feat-map}
\end{table}

\section{AAL Classifier Examples}
\label{app:class-exs}

        \begin{table}[!htbp]
            \centering
            \scriptsize
            \begin{tabular}{| P{1.5cm} | P{4.3cm} | P{0.6cm} |}
                \hline
                Corpus & Text & AAL Prob. \\
                \hline
                Dolmino (\textit{Dolmino-mix}) & \textit{...Title: heal yea Review: i b chillin wit da dawg 4 real luda iz da tytest rhyma in da hystory of da word 4 eva!! He got mad skillz cuz he keep rhymin bout hiz hoez which iz tyte cuz my whyte azz dawgs keep chillin at my plaze wit dis CD YO!! It real tyte how he spell mouf 4 real...} & 0.91  \\
                \hline
                RedPajama-v1 & \textit{...POPULAR: wiz khalifa, moneybagg yo, young jeezy, kevin gates, lil wayne, curren\$y, fabolous, nba youngboy, eminem, mac miller, meek mill, lil baby, future} & 0.80 \\
                \hline
                RedPajama-v1 & \textit{I've Changed (Interlude) [feat. Lil' Mo] 1:06..$\backslash$n Is This Our Last Time (feat. Fabolous) 5:26..$\backslash$n One Minute Man (feat. Jay-Z) [Remix] 4:36..$\backslash$n Ragtime Interlude / I'm Really Hot 3:31.. Higher Ground (Prelude) [Hidden Track] 5:02..$\backslash$n Dats What I'm Talkin About (feat. R. Kelly) 4:49..} & 0.75 \\
                \hline
                DCLM & \textit{Back to the previous page$\backslash$n Artist: Too \$hort f/ E-40$\backslash$n Album:  Pimpin' Incorporated$\backslash$n Song:   Cootie Cootie Coo$\backslash$n Typed by: OHHLA Webmaster DJ Flash...} & 0.61 \\
                \hline
                FineWeb-Edu & \textit{..."Duh wite root," pointing to a wild shrub, "dey use fuh stomach troubles. Buttuh root an palmettuh root an May apple, yuh bile tuhgedduh wid a quawt uh watuh till it simmuh down tuh haf uh pint, den yuh add some cawn wisky. Dat a fambly tonic tuh buil yuh up."...} & 0.51 \\
                \hline
            \end{tabular}
            \caption{Abbreviated examples of texts and accompanying posterior probabilities of AAL according to the demographic alignment classifier. Documents are sampled from each of 5 bins between 0.5 and 1.0 including all corpora. Texts are shown exactly as they appear in corpus documents.}
            \label{tab:class-ex}
        \end{table}

        \begin{table}[!htbp]
            \centering
            \scriptsize
            \begin{tabular}{| P{1.5cm} | P{4.3cm} | P{0.6cm} |}
                \hline
                Corpus & Text & AAL Prob. \\
                \hline
                The Pile & \textit{...<write><data>pass$\backslash\backslash$x0a</data> </write>$\backslash$n    <read><delim>$\backslash\backslash$x3e</delim><match> <pcre>.*?45 B ></pcre></match></read>$\backslash$n    <write><data>8 16$\backslash\backslash$x0a</data></write>$\backslash$n    <read><delim>$\backslash\backslash$x3e</delim><match> <pcre>.*?46 B ></pcre></match></read>....} & 0.57  \\
                \hline
                Dolmino (\textit{Olmo-Mix}) & \textit{$\# \# \# \# \# \# \# \# \# \#$ $\# \# \# \# \# \# \# \# \# \#$ $\# \# \# \# \# \# \# \# \# \#$ $\# \# \# \# \# \# \# \# \# \#$$\backslash$n$\#$$\backslash$n$\#$ Description$\backslash$n$\#$ ==================== ==================== ==================== ==================$\backslash$n$\#$$\backslash$n$\#$   General tests.....} & 0.80 \\
                \hline
                RefinedWeb & \textit{Song...$\#$$\backslash$n Yukmouth - Ooh! Ooh! Lyrics$\backslash$n Artist:$\backslash$n Yukmouth$\backslash$n Song title: Ooh! Ooh!} & 0.75 \\
                \hline
                Dolmino (\textit{Olmo-Mix}) & \textit{<reponame>Feqzz/qZoom-Client<gh_stars>1-10$\backslash$n $\#$ifndef PARTICIPANT_H$\backslash$n $\#$define PARTICIPANT_H$\backslash$n $\backslash$n $\#$include <QImage>} & 0.61 \\
                \hline
                FineWeb & \textit{Can you be hard rocking a Sade beat? Killa Kyleon will Get Rich Or Die Tryin’ to pull it off. Check it out and tell us what you think.$\backslash$n Click here to download..} & 0.51 \\
                \hline
            \end{tabular}
            \caption{Abbreviated examples of texts and accompanying posterior probabilities of AAL according to the demographic alignment classifier. Documents are sampled from each of 5 bins between 0.5 and 1.0 including all corpora.}
            \label{tab:class-ex2}
        \end{table}

    \autoref{tab:class-ex} presents randomly sampled abbreviated texts from the AAL subsets of the corpora analyzed. As shown in \autoref{tab:aal-freq}, the proportion of documents determined to contain identifiable AAL features by annotators is low for high posterior probability ranges. Considering the demographic alignment classifier \cite{blodgett-variation} uses token-level features, some of these high AAL probability documents contain repeated abbreviations while others contain dense usage of AAL as shown in the top row of \autoref{tab:class-ex}.

\section{Full Feature Analyses}
\label{app:feature-results}

We measure the distribution of AAL features across all corpora using random samples of 50,000 documents~\footnote{Some AAL subsets of corpora are smaller than 50,000. For these corpora, we consider all documents.} meeting the AAL threshold probability of 0.3. While there are some commonalities among corpora, such as Zero Plural (ZP) and Zero Copula (ZC) appearing frequently related to most other features, the frequency of individual features and order of most frequent features varies widely (e.g., Remote \textit{done} (RD) is highly frequent in OpenWebText, but not in others.)

\begin{figure*}[htbp]
    \centering
    \includegraphics[width=\linewidth]{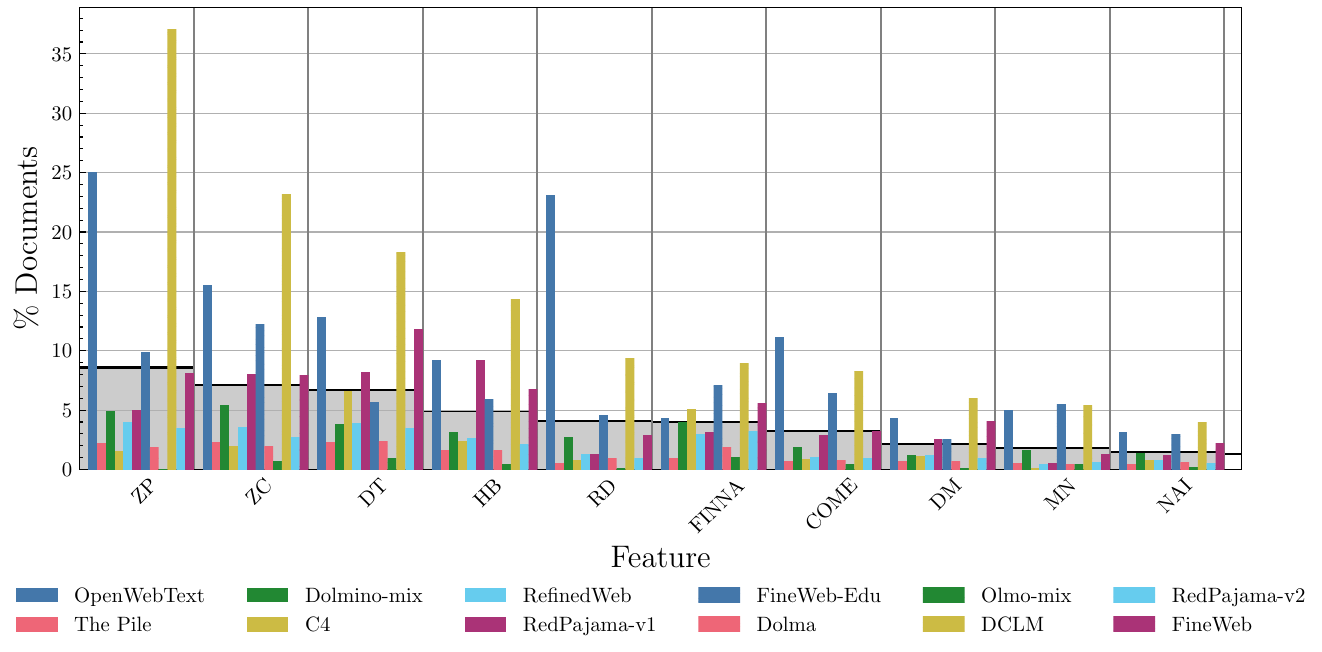}
    \caption{Feature distribution over documents for all corpora. Shaded grey region indicates the average frequency across all corpora.}
    \label{fig:all-feats}
\end{figure*}

\section{Hip Hop n-gram Analysis}
\label{app:hip-hop-method}

    To estimate the inclusion of hip hop lyrics in C4, we use infinigram to efficiently count overlapping n-grams \cite{liu2024infini}. We first strip all capitalization and punctuation from both pretraining corpus documents (C4) and hip hop lyrics\footnote{\url{https://www.kaggle.com/datasets/nikhilnayak123/5-million-song-lyrics-dataset}} to account for minor differences in the presentation of the same lyrics from different sources. We then build a suffix array of the cleaned hip hop lyrics corpus using the Llama tokenizer, 16 cpus, 512MB of RAM, 1 shard. The total build time of the suffix array was approximately 10 minutes and the resulting index is approximately 6GB.

    Throughout documents in the corpus, hip hop lyrics may be quoted by an author as part of a larger document. To account for this, we follow \cite{brown-gpt-3} and consider n-gram token lengths between 8 and 13 (inclusive) to search for overlap. We exhaustively search for all occurrences of C4 n-grams in the hip hop lyrics corpus, considering any occurrence of a documents' n-gram to be a positive label for the document as a whole. We note that this approach is not able to capture variation in orthography (e.g., spelling "going" as "goin") and relies on the comprehensiveness of the hip hop lyrics corpus; as such, it likely does not capture all hip hop lyric occurrences and underestimates actual overlap.

\section{All Corpora Hip Hop Analysis}
\label{app:all-hip-hop}

\begin{figure}
            \centering
            \includegraphics[width=\columnwidth]{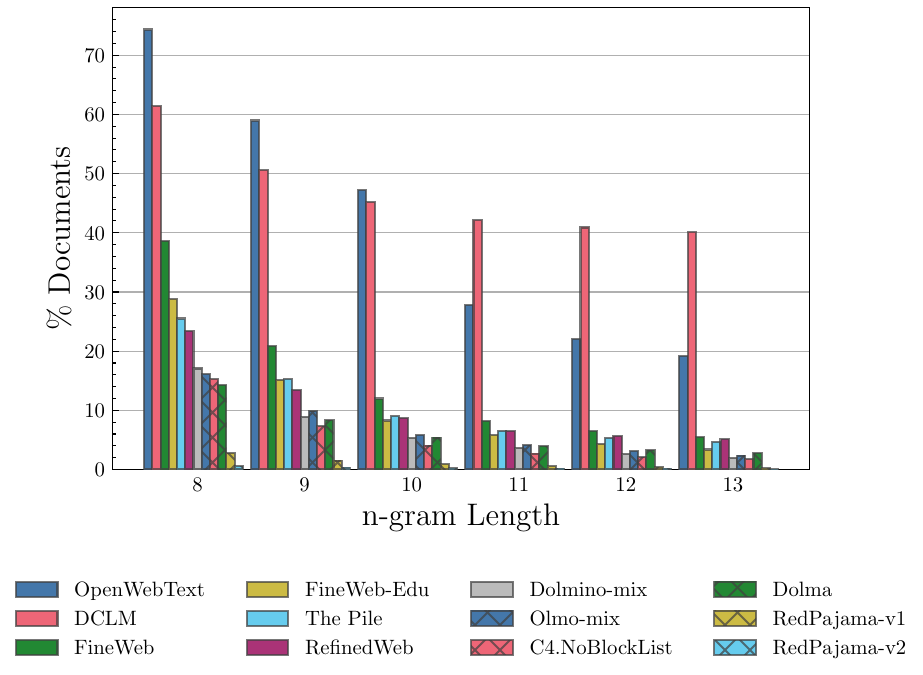}
            \caption{\% of AAL documents extracted from each corpus containing shared n-grams with hip hop lyrics for varying n-gram token lengths. Tokenization uses the Llama-2 tokenizer.}
            \label{fig:full-hiphop}
        \end{figure}

    We analyze a 50\% random sample\footnote{We consider all of the AAL texts extracted from OpenWebText given its size.} of all pretraining corpora and measure the presence of n-grams shared with hip hop lyrics. \autoref{fig:full-hiphop} presents the results of the hip hop lyrics analysis for all corpora. The proportion of documents containing hip hop n-grams varies widely across corpora, with 13-gram hip hop sequences appearing in between roughly 2-50\% of documents. Overall, however, we see that language resembling hip hop is substantially represented in each corpus. 
    
\section{Additional Hip Hop Analysis}
\label{app:add-hip-hop}

        \begin{figure}
            \centering
            \includegraphics[width=1.0\linewidth]{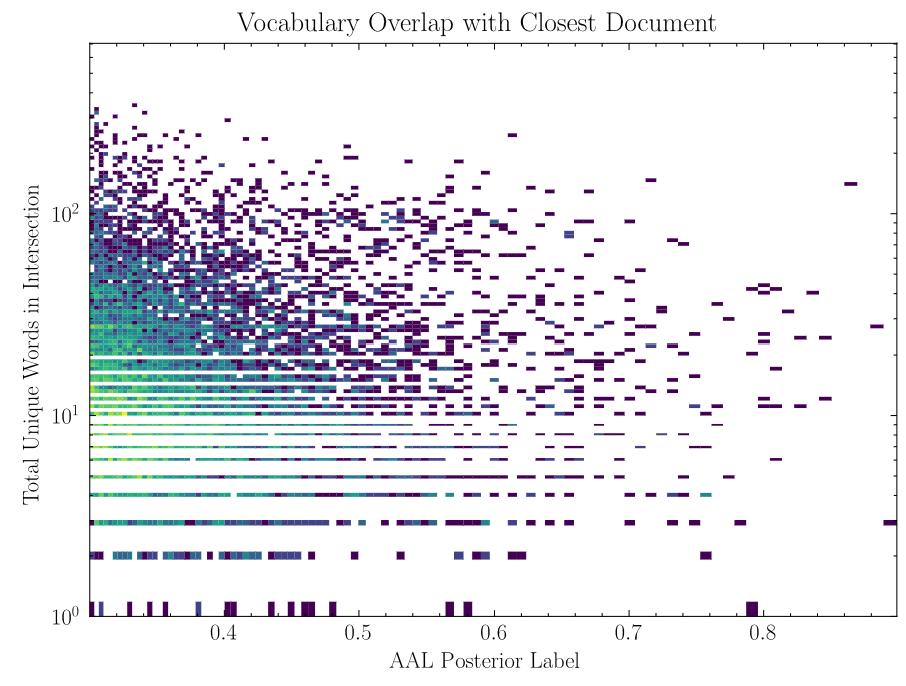}
            \caption{Plot of vocabulary overlap between AAL-labeled C4 and Rap Lyrics corpus, with 10,000 random samples. AAL Posterior Label represents the AAL probability output by the demographic alignment classifier \protect\cite{blodgett-variation}. Lighter shades indicate a higher frequency of documents. 
            }
            \label{fig:lyrics}
        \end{figure}

        To complement the hip hop lyrics n-gram overlap analysis, we additionally examine the nearest neighbors of sentences in the C4 subset with the most popular hip hop songs from a separate corpus of lyrics.\footnote{\url{https://www.kaggle.com/datasets/jamiewelsh2/rap-lyrics}} Each song and AAL document extracted from C4 is represented by a bag-of-words (BOW) vector. We then find the nearest neighbor in the hip hop lyrics corpus by extent of unique token overlap. 
        \autoref{fig:lyrics} depicts the overlap in vocabulary between rap lyrics and 10,000 randomly sampled AAL texts extracted from C4. Overall, many documents share a substantial proportion of tokens with hip hop lyrics included in the lyrics corpus. In particular, 3.20\% of sampled AAL texts exceed a threshold of 100 unique terms shared with hip hop lyrics.

\section{Non-AAL Annotations}
\label{app:non-aal}

        \begin{figure}[!htbp]
            \centering
            \includegraphics[width=1.0\linewidth]{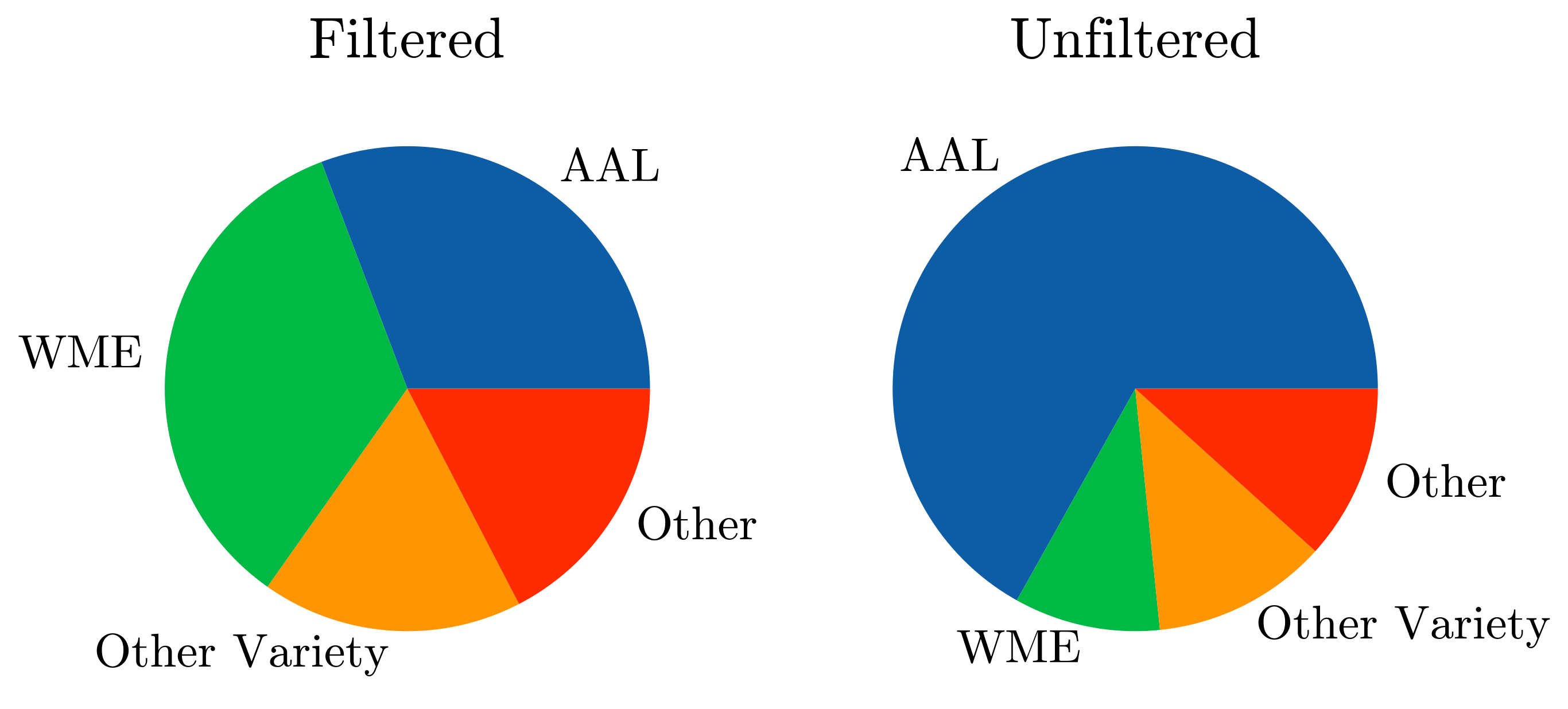}
            \caption{Proportion of texts in the filtered (\filtered{}) and unfiltered (\unfiltered{} - \filtered{}) labeled as reflecting AAL, WME, a different language variety, or no particular linguistic community.}
            \label{fig:not-aal-pie}
        \end{figure}%
        
        Given the commonalities between AAL and other varieties of English (e.g., Southern White English, \citealt{oetting-swe}), we then sought to investigate the texts that were labeled as AAL by the demographic-alignment classifier but not by annotators. \autoref{fig:not-aal-pie} presents the proportion of AAL, WME, a different variety of English, 
        and other texts (e.g., website keywords for search engines) in the filtered and unfiltered subsets of human-annotated texts. Similar to other analyses, AAL is more represented in the unfiltered proportion of texts in C4 identified as AAL by the demographic alignment classifier. Interestingly, annotators noted a significant proportion of texts in both subsets as reflecting other varieties.

\section{Automated Filters}
\label{app:filters}

\subsection{Filter Details}
    
    Descriptions of each of the \numfilter{} data filters evaluated are included in \autoref{tab:filter-dets}. 
    \textbf{Language} filters are typically neural models used to filter datasets to a particular language (e.g., English in the case of C4 and others). \textbf{Classifier-based} filters train small classification models (typically using fastText \cite{joulin-fasttext}) or n-gram models \cite{heafield-kenlm} to model text quality either through prediction of similarity to designated high-quality sources, prediction of toxicity, or n-gram model perplexity. Finally, \textbf{LLM-Based} filters leverage LLMs to directly predict aspects of data quality or annotate data to train a distilled model to evaluate quality.

    \begin{table*}[!htbp]
        \centering
        \small
        \begin{tabular}{| c |c P{6cm}|}
            \hline
            Filter & Example Use & Description \\
            \hline
            fastText-langid \cite{lui-langid} & RefinedWeb & fastText classifier trained to detect English text \\
            fastText OH-2.5 + ELI5 \cite{datacomp-lm} & DataComp-LM* & fastText model trained on the OpenHermes and ELI5 datasets \\
            OpenWebText2 vs. RefinedWeb \cite{datacomp-lm} & DataComp-LM* & fastText classifier trained with OpenWebText2 as the positive class and RefinedWeb as the negative class \\
            Wikipedia vs. RefinedWeb \cite{datacomp-lm} & DataComp-LM* & fastText classifier trained with Wikipedia as the positive class and RefinedWeb as the negative class \\
            Multi-source vs. RefinedWeb \cite{datacomp-lm} & DataComp-LM* & fastText classifier trained with Wikipedia, RedPajama, Books, and OpenWebText2 as the positive class and RefinedWeb as the negative class \\
            Perplexity Filtering \cite{datacomp-lm} & DataComp-LM* & Wikipedia-trained ngram model \\
            fastText Toxicity \cite{soldaini-dolma} & Dolma & fastText classifier trained on the Jigsaw toxicity dataset \\
            PALM Quality Filter$^\dagger$ \cite{chowdhery-palm} & PALM & fastText classifier trained to distinguish Wikipedia, OpenWebText, and RedPajama-v1 Books as the positive class from Common Crawl. \\
            Wikipedia-like Classifier \cite{redpajama} & RedPajama-v1 & fastText classifier trained to distinguish Wikipedia-like text from others \\
            CC Quality Filter$^\dagger$ \cite{thepile} & The Pile & fastText classifier trained to distinguish high-quality data sources and Common Crawl in general \\
            FineWeb-Edu \cite{fineweb} & FineWeb-Edu & \texttt{Snowflake-arctic-embed}-based classifier trained on \texttt{Llama-3-70b-Instruct} annotations of educational content \\
            AskLLM \cite{sachdeva-askllm} & DataComp-LM* & Prompts an LLM to measure pretraining data quality \\
            QURating \cite{wettig-qurating} & QURating & Prompts an LLM with 4 criteria to judge pretraining data quality \\
            \hline
        \end{tabular}
        \caption{List of the \numfilter{} evaluated pretraining corpora filters, including an example dataset or study employing each filter and a brief description. * denotes that the filter was evaluated in the development of the cited dataset, but not used in a resulting corpora. $^\dagger$ indicates filters that were replicated; all other filters are taken from their respective cited sources.}
        \label{tab:filter-dets}
    \end{table*}

\subsection{Replicated Filters}

    Where possible, we use filters provided by the papers that use or evaluate them. For two filters, the PALM quality filter and CC quality filter, we replicate them following available descriptions. For both filters, we train supervised fastText classifiers \cite{joulin-fasttext} with the default hyperparameters.

    \textbf{PALM Quality Filter. } The PALM quality filter \cite{chowdhery-palm} uses Wikipedia samples, OpenWebText, and RedPajama-v1 books as the positive class and Common Crawl as the negative class. Because the books subset of RedPajama-v1 is unavailable, we instead use the Project Gutenberg books corpus.~\footnote{\url{https://huggingface.co/datasets/manu/project_gutenberg}} We use publicly available corpora for Wikipedia~\footnote{\url{https://huggingface.co/datasets/wikimedia/wikipedia}} and OpenWebText~\footnote{\url{https://huggingface.co/datasets/Skylion007/openwebtext}} \cite{gokaslan-openwebtext}. We sample 600 CCNet snapshots weighted among head (10\%), middle (20\%), and tail (70\%) partitions made available through and following RedPajama-v2 \cite{redpajama-v2}. We sample 250,000 positive and negative references, with positive references split evenly among the three corpora.

    \textbf{Pile-CC.} The Pile-CC quality filter \cite{thepile} uses OpenWebText \cite{gokaslan-openwebtext} as the positive class and Common Crawl as the negative class. Similarly to the PALM filter, we simply sample 250,000 OpenWebText and CCNet samples to form the training data.

\section{AAL Sources in Pretraining Corpora}
\label{app:aal-urls}

\begin{figure}
    \centering
    \includegraphics[width=0.95\linewidth]{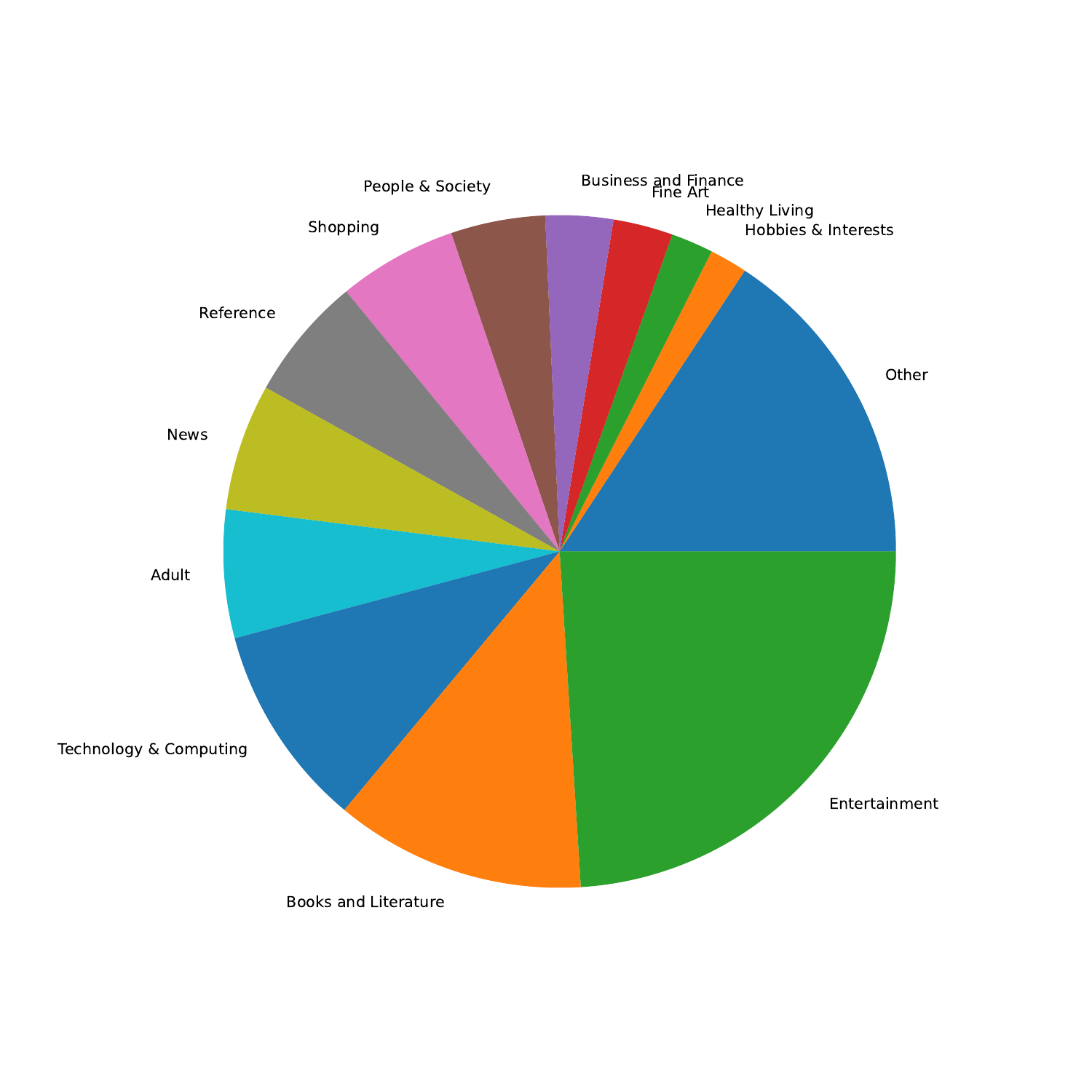}
    \caption{Distribution of AAL source URL IAB categories.}
    \label{fig:url-dist}
\end{figure}

We analyze the sources of AAL texts by examining the associated URL categories according to the IAB taxonomy.~\footnote{\url{https://www.iab.com/guidelines/content-taxonomy/}} Using the URL's listed in the C4 corpus, we cross reference the network location of each with a dataset of 99,015 popular website IAB classifications.\footnote{\url{https://www.kaggle.com/datasets/bpmtips/websiteiabcategorization}} A small proportion of texts (~5\%) are associated with URLs that are included in the dataset, suggesting that many AAL texts are not sourced from highly-visited sites. Among texts that could be matched to URL classifications, \autoref{fig:url-dist} shows the distribution of IAB categories. A majority of texts come from \textit{Entertainment} domains, containing content such as hip hop lyrics, while large proportions are also drawn from \textit{Technology \& Computing} domains such as social media and other forums.

\end{document}